\documentclass{article}
\usepackage{arxiv}

\usepackage[utf8]{inputenc} 
\usepackage[
    colorlinks=true,
    linkcolor=blue,
    citecolor=blue,
    urlcolor=blue
]{hyperref}

\usepackage[T1]{fontenc}
\usepackage{graphicx} 
\usepackage{algorithmic}
\usepackage{amsmath}
\usepackage{color}
\usepackage{subcaption}  
\usepackage{amsmath} 
\usepackage{xcolor}
\usepackage{booktabs}
\usepackage{url}
\usepackage{pdflscape}
\usepackage{float}
\usepackage{multirow}
\usepackage{booktabs}
\usepackage{array}
\usepackage{algorithm} 
\usepackage[table]{xcolor}
\usepackage{subcaption}
\usepackage{adjustbox}
\usepackage{longtable}
\usepackage{rotating} 

\usepackage{subcaption}
\newcommand{\ignore}[1]{}

\title{Quality Diversity for Reliable Data Driven Time-Use Optimization}

\begin{document}

\author{{Aneta Neumann} \\
 Optimisation and Logistics\\
	School of Computer and Mathematical Sciences\\
        Adelaide University\\
        Adelaide, Australia \\
        \And
{Ty Stanford} \\
	 School of Allied Health and Human Performance\\
        Adelaide University\\
        Adelaide, Australia \\
        \And
{Dorothea Dumuid} \\
	 School of Allied Health and Human Performance\\
        Adelaide University\\
        Adelaide, Australia \\
        \And
{Frank Neumann} \\
 Optimisation and Logistics\\
	School of Computer and Mathematical Sciences\\
        Adelaide University\\
        Adelaide, Australia \\
        }

\maketitle

\begin{abstract}
The daily allocation of the finite 24-hour time budget is strongly associated with physical, mental, and cognitive health. While predictive models can estimate the relationship between time-use compositions and health outcomes such as body mass index, life satisfaction, and cognition, most optimization approaches focus only on maximizing expected benefit and do not consider the uncertainty inherent in data-driven prediction. Ignoring uncertainty in health-related decisions can lead to unrealistic time-use recommendations. To address this gap, we introduce an uncertainty quantification Quality Diversity (QD) framework for a more reliable time-use recommendation. Objective functions are derived using compositional data analysis using a large child cohort dataset $n > 1000$, to capture the relationship between daily activity compositions and multiple health indicators. We develop a new approach that incorporates predictive uncertainty into QD processes and produces more reliable recommendations that balance the expected health benefits with the confidence of the model. We explore the solution space through variable-based and objective-based behavioral representations, revealing diverse high-quality time-use composition and explicit relationships between health outcomes under uncertainty. By embedding uncertainty directly into optimization, our framework shifts the time-use recommendations toward regions of lower uncertainty while preserving high-quality structures for more reliable decision-making in behavioral health.
\end{abstract}


\section{Introduction}
\label{sec:intro}
Time-use optimization problems aim to provide guidance on how to structure daily activities in order to improve health outcomes. These problems are formulated using predictive models trained on real-world datasets, which estimate relationships between time-use compositions and outcomes such as physical, cognitive, and psychological health. However, such models are inherently uncertain due to limited data coverage, measurement noise, and extrapolation effects in compositional spaces.
Traditional time-use optimization approaches focus on maximizing expected health benefits but thereby overlook the uncertainty associated with these predictions~\cite{DBLP:conf/ppsn/XieNSRDN22,nikfarjam2024quality}. 
By explicitly incorporating uncertainty into the optimization framework, recommendations can shift towards solutions that balance expected benefits with reliability, revealing more robust strategies. This probabilistic perspective improves the interpretability of results and provides insights that standard approaches may miss, enabling more informed decision-making in health-related contexts.
In this paper, we develop a Quality Diversity framework that explicitly considers uncertainty in the high-dimensional behavioral space for time-use optimization, to make reliable data-driven recommendations.
\subsection{Background and related work}
Daily time use is a constrained 24-hour constant-sum budget where activities are mutually exclusive and exhaustive. This co-dependency has driven a paradigm shift in behavioral epidemiology toward Compositional Data Analysis (CoDA), which evaluates health outcomes based on time reallocations rather than isolated behaviors~\cite{Dumuid2018,Dumuid2022a}. A systematic review of studies using compositional data
analysis~\cite{miatke2023} confirmed that these daily compositions significantly affect adiposity, cardiometabolic health, and mental wellbeing, leading global health authorities to establish integrated time-allocation guidelines. 
While the 2019 Australian 24-Hour Movement Guidelines~\cite{australian2019guidelines}, provide specific benchmarks such as 9–11 hours of sleep and $\ge 1$ hour of MVPA for children, the evidence base remains limited. 
While Quality Diversity (QD) approaches can estimate how time-use compositions relate to outcomes such as body mass index, life satisfaction, and cognition, existing optimization approaches focus on maximizing benefit and ignore the uncertainty inherent in data-driven recommendations~\cite{nikfarjam2024quality}. Exact methods identify optimal compositions for health outcomes but are computationally intensive for higher-dimensional, real-world time-use patterns and assume a single optimum, limiting choice. To address these issues, we propose an uncertainty-quantified Quality Diversity framework that enables more reliable time-use recommendations. By evaluating the confidence intervals of solutions relative to the training data, we shift the objective from simple maximization to a reliability-based optimization, ensuring that the recommended time-use compositions are high-performing and statistically robust.

Quality-Diversity (QD) algorithms, which have gained considerable momentum in recent years, aim to find a set of high-performing solutions that are explicitly diverse with respect to behavioral descriptors~\cite{DBLP:conf/cec/LehmanS11,pugh2016quality,QIN2026102240,math14071091}. 
QD introduces mechanisms to preserve elite individuals while encouraging exploration of sparsely and noisy populated behavioral regions~\cite{DBLP:conf/gecco/CullyM13,DBLP:conf/gecco/GrillottiFLC23,DBLP:conf/gecco/FlageatHHDC25}. MAP-Elites, introduced by~\cite{DBLP:journals/corr/MouretC15}, discretize a behavioral space into cells and maintain the best-performing solution per cell, and enable illumination of the performance landscape rather than convergence to a single optimum. This competition allows QD algorithms to evolve a diverse, high-quality behavioral set of solutions that are particularly valuable in complex and high-dimensional domains. 
QD has been successfully applied in robotics~\cite{DBLP:journals/nature/CullyCTM15,DBLP:journals/firai/NordmoenVEG21,DBLP:conf/gecco/MertanC25}, logistics~\cite{10.1145/3712256.3726343}, communication networks~\cite{DBLP:conf/gecco/Gounder0N24,DBLP:conf/gecco/NeumannGYSCG023}, control systems and game design~\cite{DBLP:journals/tciaig/AlvarezDFT22,DBLP:conf/gecco/MedinaRMS23,DBLP:conf/eurogp/ZhangCTXBZ23}.
More recently, the paradigm has been applied in combinatorial optimization such as traveling thief problems~\cite{nikfarjam2024use}, coevolutionary Pareto optimization~\cite{DBLP:conf/gecco/NeumannA022} and monotone submodular functions~\cite{schmidbauer2024guiding,DBLP:conf/gecco/NeumannB021}.
QD was applied to behavioral epidemiology for time-use optimization to improve health outcomes~\cite{nikfarjam2024quality}. The authors demonstrated that QD can navigate the constrained simplex of 24-hour time-use to identify a diverse set of optimal health-promoting compositions. While QD has shown significant promise in behavioral modeling, its capacity for uncertainty quantification within the context of time-use recommendations remains largely underexplored.
\subsection{Our contribution}
In this paper, we introduce a data driven time-use optimization framework based on Quality Diversity (QD) that allows systematic exploration of solutions not only in terms of health benefits  but also the reliability of diverse behavioral recommendations. Unlike standard approaches that optimize expected outcomes alone, our framework explicitly accounts for uncertainty arising from data-driven prediction models.
We show that ignoring uncertainty leads to the selection of solutions in weakly supported regions of the model, which can result in unreliable or unrealistic time-use recommendations. To address this, we incorporate uncertainty directly into the QD process, propose methods to reduce uncertainty through reliable data selection and discounting, and provide systematic analysis of uncertainty in behavioral QD maps.

We develop our framework in the context of health-related time-use optimization, where daily activity compositions are linked to outcomes such as physical, psychological, and cognitive health using models derived from a large cohort dataset. Solutions correspond to 24-hour activity allocations, with predicted benefits and associated uncertainties computed from these models.
We investigate variable-based and objective-based behavioral spaces, revealing diverse sets of high-quality solutions and highlighting how uncertainty shapes the structure of the solution space. Furthermore, we provide a comprehensive investigation of real-world data on multiple health objectives and analyze how uncertainty improves the reliability of recommendations.
Our framework provides a new approach for generating diverse, high-quality, and reliable time-use recommendations, offering practical value to health professionals by explicitly accounting for model confidence in decision-making. We show that uncertainty is not uniformly distributed across the solution space and can significantly influence which solutions are high-quality. Our results demonstrate that incorporating uncertainty leads to more reliable and interpretable recommendations while preserving behavioral diversity.

The paper is organized as follows. Section~\ref{sec:2} introduces Quality Diversity for time-use optimization. Sections~\ref{sec3} and~\ref{sec4} present uncertainty quantification for time-use recommendations and the QD approach for more reliable guidance. Section~\ref{sec5} carries out comprehensive experimental investigations, including $4D$ and $7D$ problems for real-world datasets. The paper concludes with final remarks.

\section{Quality Diversity for Time-Use Optimization}
\label{sec:2}

Quality Diversity algorithms, e.g., MAP-Elites, are optimization methods designed to identify high-performing "elite" solutions across a diverse behavioral spectrum~\cite{DBLP:journals/corr/MouretC15}. Unlike Evolutionary Algorithms that may converge on a single global optimum, MAP-Elites employs a niche-based survival selection. Competition is restricted to individuals with similar Behavioral Descriptors (BDs). This ensures that the archive maintains a diverse map of top-performance solutions throughout the feasible space.

We define two different settings for MAP-Elites such as Variable-Based Behavioral Spaces (VBS) and Objective-Based Behavioral Spaces (OBS) as done in~\cite{nikfarjam2024quality}.
MAP-Elites operates by mapping elites across a discretized behavioral space, where solutions compete only within regions defined by similar behavioral descriptors (BDs), thereby promoting diversity while preserving quality. In a VBS, the behavioral descriptors (BDs) are derived directly from the solution space variables ($x_i, x_j$), with the grid resolution determined by the parameter $\alpha_i$ to control cell granularity across the feasible range $[x_{min}, x_{max}]$. Each cell stores the best-performing solution associated with that region, ensuring coverage across the range of variable values. In contrast, an OBS utilizes objective function outputs ($\hat{y}_i, \hat{y}_j$) as BDs, mapping the distribution of elite solutions across the performance. Both configurations require discretizations where the number of niches is a function of the range and the user-defined resolution ($\delta_i$ for OBS), ensuring that survival selection occurs only between solutions with localized phenotypic similarities. This formulation enables the algorithm to illuminate tradeoffs among objectives by preserving elites across different regions of the objective space. Together, these representations highlight how MAP-Elites balances exploration and exploitation by maintaining a structured archive of diverse and high-quality solutions.

\section{Data-Driven Time-Use Recommendations and Uncertainty Quantification}
\label{sec3}

In the following, we describe how we predict health outcomes based on given datasets and describe how to measure uncertainty based on the used prediction model.
For a given time-use solution $x$, we derive the expected health benefit $E(x)$ as well as the uncertainty $U(x)$ of $x$. Our model is based on multiple linear regression, but the QD approach introduced in Section~\ref{sec4} can be used for any model for which $E(x)$ and $U(x)$ can be derived.

\subsection{Observed data used in data-driven time-use predictions}

Given a dataset containing $n$ observations of an outcome (fitness)\
$$
\mathbf{y} = (y_{1}, \ldots, y_{n})^{{T}},
$$ and $n$ corresponding sets of predictor values

$$
X =
\begin{bmatrix}
1 & z_{11} & z_{12} & \ldots & z_{1D'} & z_{11}^2 & z_{11}z_{12} & \ldots & z_{1D'}^2 \\
1 & z_{21} & z_{22} & \ldots & z_{2D'} & z_{21}^2 & z_{21}z_{22} & \ldots & z_{2D'}^2 \\
\vdots & \vdots & \vdots &  & \vdots & \vdots & \vdots &  & \vdots \\
1 & z_{n1} & z_{n2} & \ldots & z_{nD'} & z_{n1}^2 & z_{n1}z_{n2} & \ldots & z_{nD'}^2
\end{bmatrix},
$$ 

where $\mathbf{x}_{i} = \left( x_{i1},x_{i2},\ldots, x_{iD}  \right)^T$ is the $D \times 1$ time‑use vector of person $i=1,2,\ldots,n$ with constraint $\sum_{j=1}^{D} x_{ij} = 1440$ minutes, ${\mathbf{z}}_{i} = \textit{ilr} (\mathbf{x}_{i}) = \mathbf{V}^T \,\ln_*(\mathbf{x}_{i})$ with $\mathbf{V}^{{T}}$ a $D' \times D$ a valid orthonormal basis matrix for the chosen isometric log-ratio (\textit{ilr}) transform with $D' = D - 1$, and $\ln_*({.})$ is an element-wise operator of the natural logarithm. Note that the terms $z_{i1}^2, z_{i1}z_{i2},  \ldots, z_{iD'}^2$ allow the fitness function to vary non-linearly with the \textit{ilr}s, if statistically significant (otherwise are removed).

\subsection{Multiple linear regression model}

A multiple linear regression model fit on these data to linearly relate the observed outcomes, $\boldsymbol{y}$, from the predictor values is written
$$
\boldsymbol{y} = X\boldsymbol{\beta} + \boldsymbol{\epsilon}
$$
where $\boldsymbol{\beta}$ is a vector of $D' + D'(D'-1)/2 + 1$ coefficients and $\boldsymbol{\epsilon} \sim \mathcal{N}_n(\boldsymbol{0}_n, \sigma^2 I_n)$ is the model error in the predicted outcomes, $X\beta$, from the observed outcomes, $y$.

Multiple linear regression uses the method of least-squares, which minimises the summed squared differences between the observed outcomes $\boldsymbol{y}$ and predicted values $X\boldsymbol{\beta}$, written 
\[ 
\arg\min_{{\boldsymbol{\beta}} } \| \boldsymbol{y} - X {\boldsymbol{\beta}}  \|^2.
\]

If the columns of $X$ are linearly independent,
$
\hat{\boldsymbol{\beta}} = (X^T X)^{-1} X^T \boldsymbol{y}
$
uniquely minimizes the summed squared differences. This is the reason the \textit{ilr} transformation of the time-use variables as the constraint $\sum_{j=1}^{D} x_{ij} = 1440$ minutes necessarily makes the time-use variables linearly dependent (which can easily be seen as $x_{iD} =  1440 - x_{i1}- x_{i2}- \ldots - x_{iD'}$ for every person $i=1,2,\ldots, n$). 
Like all statistical models, there are assumptions that need to be met (namely, independent observations and $\boldsymbol{\epsilon} \sim \mathcal{N}_n(\boldsymbol{0}_n, \sigma^2 I_n)$). These assumptions can be assessed using the experimental design (independence of observations) and diagnostic plots of the residuals ($y - X \hat{\boldsymbol{\beta}}$, residual normality with constant variance and no systematic relationships with the predictors).

\subsection{Prediction on a new observation}
Using the assumed statistical model, the predicted fitness for a new set of predictor values $\boldsymbol{x}_0$ is written
$
{Y}_0 = \boldsymbol{x}_0^T \boldsymbol{\beta} + {\epsilon}_0.
$
As ${\epsilon}_0$ is distributed around a mean of 0, the expected value of ${Y}_0$ is equal to $\boldsymbol{x}_0^T \boldsymbol{\beta}$. However, as we do not know the true population parameter vector of $\boldsymbol{\beta}$, we instead must use the sample-estimated predicted value of $\boldsymbol{x}_0^T \hat{\boldsymbol{\beta}}$ so we define
$
{E}^*(\boldsymbol{x}_0)   = \boldsymbol{x}_0^{T} \hat{\boldsymbol{\beta}} 
$
(where $E^*(.)$ is the prediction function in this context and not to be confused with the expectation of a random variable or random vector).
As the function $E^*(.)$ expects a full regression set of predictors (e.g., all the elements in a row of the "design matrix" $X$ previously denoted as $\boldsymbol{x}_0$), we will define a convenience predictive function we will use herein as
$$
{E}(\boldsymbol{x}) := {E}^*(\boldsymbol{x}_0)
$$
where 
$
\boldsymbol{x}_0^T  
= \left[ 1 \,\,\, z_1 \,\,\, z_2 \, \ldots \, z_{D'} \,\,\, z_1 z_1 \,\,\, z_1z_2 \,\,\, \ldots \,\,\, z_{D'}z_{D'} \right]
$
so that only the time-use composition vector $\boldsymbol{x}$ needs to be provided to the prediction function ${E}(\boldsymbol{x})$.

\subsection{Uncertainty quantification for a new observation}

Once again, for a new set of predictors, $\boldsymbol{x}_0$, the assumed model is
$
{Y}_0 = \boldsymbol{x}_0^T \boldsymbol{\beta} + {\epsilon}_0
$
so the prediction error variance can be re-expressed as
$
\mathrm{var}( {\epsilon}_0) = \mathrm{var}({{Y}}_0 - \boldsymbol{x}_0^{T} {\boldsymbol{\beta}}  ).
$
However, like before, we must use the sample-based estimates instead of the unknown population parameter vector, ${\boldsymbol{\beta}}$, so we use the estimated variance of the error, noting that the variance of the estimated parameter vector is $\mathrm{var}( \hat{\boldsymbol{\beta}} ) = \sigma^2 (X^{T} X)^{-1}$,

$$
\begin{aligned}
\hat{\mathrm{var}}({\epsilon}_0) 
&= 
\mathrm{var}(Y_0 - \boldsymbol{x}_0^{T} {\hat{\boldsymbol{\beta}}}  ) \\
&=
\mathrm{var}(Y_0) + 
\boldsymbol{x}_0^{T} \, 
\mathrm{var}( \hat{\boldsymbol{\beta}}) \, 
\boldsymbol{x}_0  
\text{ as $Y_0$ and $\hat{\beta}$ are independent} \\
&=
\sigma^2 + \sigma^2 \boldsymbol{x}_0^{T} \, (X^{T} X)^{-1}  \, \boldsymbol{x}_0 
\text{ as } Y_0 \sim N(0, \sigma^2)  \\
&=
\sigma^2 \left( 1 +  \boldsymbol{x}_0^{T} \, (X^{T} X)^{-1}  \, \boldsymbol{x}_0 \right) .
\end{aligned}
$$

From the above, we can see the uncertainty of a prediction ${E}^*(\boldsymbol{x}_0)$ is a function of (a) the variance in the outcome (because of the normally distributed model error), and (b) the variance in the sample-based prediction $\boldsymbol{x}_0^{T} \hat{\boldsymbol{\beta}}$ (as opposed to using the true population parameter vector $\boldsymbol{\beta}$). Therefore we denote the uncertainty in a predicted outcome as,
$$
\begin{aligned}
U^*\left(\boldsymbol{x}_0\right)
=\hat{\mathrm{SD}}(E^* \left(\boldsymbol{x}_0\right))
= \sqrt{\hat{\mathrm{var}}(E^* \left(\boldsymbol{x}_0\right) )}
  = \sigma \sqrt{1 +\boldsymbol{x}_0^{T} (X^{T} X)^{-1} \boldsymbol{x}_0 }.
\end{aligned}
$$

Similarly to the convenience function in prediction, ${E}(\boldsymbol{x}) := {E}^*(\boldsymbol{x}_0)$, we define a convenience uncertainty function used herein as
$$
{U}(\boldsymbol{x}) := {U}^*(\boldsymbol{x}_0)
$$
where 
$\boldsymbol{x}_0^T =
 \left[ 1 \,\,\, z_1 \,\,\, z_2 \, \ldots \, z_{D'} \,\,\, z_1 z_1 \,\,\, z_1z_2 \,\,\, \ldots \,\,\, z_{D'}z_{D'} \right]$
as before.

Note that $U^*(\boldsymbol{x}_0)$ does not depend on the predicted value $E^*\left(\boldsymbol{x}_0\right) = \boldsymbol{x}_0^{T} \hat{\boldsymbol{\beta}}$ itself but only it varies only with inputs (solutions) $\boldsymbol{x}_0$, as $\sigma$ (estimated by $\hat{\sigma} = s_e = \frac{1}{n - p} \|Y - X\hat{\beta}\|$) and the design matrix $X$ are constant for any (new) input.
The variance (and therefore the standard deviation/uncertainty) of a prediction increases as $\boldsymbol{x}_0^{T}$ moves away from the column‑wise mean of the predictor variables
$
\mu_X^{T} = \mathrm{mean}_{\mathrm{columnwise}}(X)
$
relative to the estimated predictor variable variance (and covariances) as seen in the term, $\sigma^2 \, \boldsymbol{x}_0^{T} (X^{T} X)^{-1} \boldsymbol{x}_0$.

\section{QD Approach for More Reliable Recommendations}
\label{sec4}

In the following, we assume that we maximize given function $E(x)$ measuring the health outcome of time-use.
Based on the given set of data, we have developed a model that allows to compute for each given time-use the solution the expected health benefit $E(x)$ and uncertainty $U(x)$. The optimization of the expected health benefit using QD has been considered in \cite{nikfarjam2024quality} and we study an extension of this approach to incorporate uncertainty. The first part is to observe with each of the optimized solutions computed in the behavioral space not only the expected health benefit $E(x)$ but also its uncertainty $U(x)$. Observing $U(x)$ in addition to $E(x)$ allows the user to judge how reliable to different solutions provided by QD in the map are and identify regions of reliable high quality solutions. If the uncertainty for solutions with a high expected health benefit is low, then this provides additional confidence in the time-use recommendation achieving the highest expected benefit. If the uncertainty is high for solution of high expected benefit, then the user might shift its focus to solutions with a lower uncertainty but also slightly lower expected health benefits in order to reduce the risk associated with the considered solution.

We use fitness functions of the form
$
F(x) = E(x) - C_{\alpha} \cdot U(x)
$
which should be maximized.
Here, $E(x)$ is the expected score for solution $x$ according to our data model and $U(x)$ is the uncertainty of the prediction for solutions $x$ with respect to the given data set as derived in Section~\ref{sec3}. 
In the case, where we aim to minimize the health outcome, we use
fitness functions of the form
$
F(x) = E(x) + C_{\alpha} \cdot U(x)
$
which should be minimized.

To define a different certainty level, we use $C_{\alpha}$ based on the $\alpha$-fractile point of the standard normal distribution.
We note that uncertainty is not necessarily normally distributed in this way. 
We use $C_{\alpha}=6.36134$ which is motivated by the $\alpha$-fractile point of the standard normal distribution for
$\alpha=1 -10^{-10}$. We note that differences in expected benefit are usually much higher than differences in uncertainty quantification. This means that a factor of 
 $C_{\alpha}=6.36134$ has a relatively moderate effect on the shift in expected health benefit.
We also consider the case $C_{\alpha}=0$ which means that only the expected value $E(x)$  is used to direct the search. Using $C_{\alpha}=0$, we call it the standard model and it  matches the setting considered in~\cite{nikfarjam2024use}.
For all choices of $C_{\alpha}$, we store the solutions $x$ and the values $F(x), E(x)$, $U(x)$ as part of our quality diversity approach. We note that the search is directed by $F(x)$ where as $E(x)$ and $U(x)$ for the solutions obtained are displayed in the maps obtained in our experimental investigations. 

\subsection{Quality Diversity approach}
Our algorithm introduces a reliability-driven Quality Diversity framework that departs from standard MAP-Elites for time-use optimization introduced in~\cite{nikfarjam2024quality} by embedding predictive modeling and uncertainty quantification directly into the search and selection process. Algorithm~\ref{alg:qd} provides an outline for the MAP-Elites for reliable time-use recommendations in the case where a given health outcome objective should be maximized. Crossover and mutation operators to create new solutions are the same as introduced in~\cite{nikfarjam2024quality}.

 Our new approach leverages a surrogate model $M$ to estimate the expected performance $E(x)$ of candidate solutions $x$, in conjunction with an uncertainty measure $U(x)$ that captures the confidence of these predictions. This formulation enables the definition of a risk-aware fitness function, $F(x) = E(x) - C_{\alpha} \cdot U(x)$, which explicitly penalizes uncertain solutions and promotes robust decision-making. As such, the algorithm is designed not only to explore diverse regions of the behavioral space but also to prioritize solutions that are both high-performing and reliable under model uncertainty.

The search process follows a MAP-Elites structure, where candidate solutions are generated either through random sampling or variation of previously discovered elites, and mapped into a discretized behavioral space via their descriptors~\cite{nikfarjam2024quality}. 
However, selection within each cell is guided by the uncertainty-aware fitness rather than only performance. A solution replaces the current elite in a cell only if it offers a superior trade-off between predicted quality and uncertainty. This mechanism ensures that the archive evolves towards a set of solutions that are robust to prediction errors, addressing a key limitation of conventional QD approaches in data-driven contexts. The final output is a structured map that captures diverse behavioral patterns while systematically prioritizing robustness and reliability in their predicted outcomes.
\begin{algorithm}[t]
\caption{MAP-Elites for Reliable Time-Use Recommendations 
}
\label{alg:qd}
\begin{algorithmic}[1]
\STATE \textbf{Input:} model $\mathcal{M}$, empty maps $\mathcal{X} \leftarrow \emptyset, \mathcal{F} \leftarrow \emptyset, \mathcal{E} \leftarrow \emptyset, \mathcal{U} \leftarrow \emptyset$, parameters $C_{\alpha}$, $I$ (total number of fitness evaluations), $G$ (number of initial solutions). 
\FOR{$k = 1 \to I$} 
    \IF{$k \leq G$} 
      \STATE Choose solution $\mathbf{x'}$ randomly. 
    \ELSE
        \STATE Produce new solution $\mathbf{x'}$ by mutation or crossover.
        (parent(s) are selected uniformly at random from $\mathcal{X}$)
    \ENDIF
    \STATE $\mathbf{b}' \leftarrow \text{descriptor}(\mathbf{x}')$ (obtains cell corresponding to $\mathbf{x}'$ in the behavioral space)  
    \STATE Compute $E(\mathbf{x'})$ and $U(\mathbf{x'})$ based on the given model $M$
    \STATE $F(\mathbf{x}') \gets E(\mathbf{x}') - C_\alpha \cdot U(\mathbf{x}')$ 
    \IF{$\mathcal{X}(\mathbf{b}') = \emptyset$ \textbf{OR }$\mathcal{F}(\mathbf{b}') < F(\mathbf{x'})$}
        \STATE 
        $\mathcal{F}(\mathbf{b}') \leftarrow F(\mathbf{x}')$, 
        $\mathcal{E}(\mathbf{b}') \leftarrow {E}(\mathbf{x}')$, 
        $\mathcal{U}(\mathbf{b}') \leftarrow {U}(\mathbf{x}')$,     
        $\mathcal{X}(\mathbf{b}') \leftarrow \mathbf{x}'$ 
    \ENDIF
\ENDFOR
\RETURN{
$(\mathcal{X, F, E, U})$}
\end{algorithmic}
\end{algorithm}

\section{Detailed Experimental Investigations for Health Models}
\label{sec5}

This section presents a comprehensive experimental analysis of the QD algorithm under uncertainty, their performance, and the structure of behavioral spaces. First, we describe the real-world data sources and the experimental setting. We evaluate the $4D$ and $7D$ problem from~\cite{nikfarjam2024quality}, and conduct a comprehensive examination of the results at a given confidence level for several objectives.

This work utilizes data from a large population-based child cohort to demonstrate the practical application of our QD time-use approach for more reliable recommendations. Data were obtained from the Child Health CheckPoint - Growing Up in Australia study~\cite{MCRI2025}, a cross-sectional module embedded in Release 9.1 C1 (Waves 1-9) of the Longitudinal Study of Australian Children (LSAC)~\cite{BAA3N6_2021}. Children from the LSAC birth cohort, which began in 2003 with over $10,000$ participants, who remained in the study through Wave $9$ ($n = 7,658$) were invited to participate in the Child Health CheckPoint at ages $11$–$12$ years. The daily behaviors were quantified via 7-day, 24-hour wrist-worn accelerometry (4D composition: sleep, sedentary, LPA, MVPA)
and 24-hour recalls ($7D$ composition including domains e.g., screen time and school activities). 

To map these behaviors to multi-dimensional health objectives, we assessed metrics including NAPLAN academic performance, NIH Toolbox cognition, VO2 max via cycle ergometry, and standardized BMI z-scores. To handle the intrinsic collinearity and constant-sum constraint of the 24-hour budget, we employed Compositional Data Analysis (CoDA) and Simplexity using the compositions package in R~\cite{Stanford2022codaredistlm,Stanford2024simplexity}. Specifically, we addressed the singularity of the covariance matrix by applying a log-ratio expectation maximization algorithm for zero-replacement, followed by an isometric log-ratio (ilr) transformation. This mapped the $D$-part simplex into $D-1$ linearly independent coordinates, enabling the robust estimation of objective functions via least-squares regression without violating the assumption of predictor independence.

\subsection{The 4D problem}

We explore the behavioral space of the $4D$ problem under uncertainty and compare the results obtained from the discounted QD algorithm with those from a standard approach. We consider four dimensions such as sleep (SL), sedentary activities (SED), light physical activity (LPA), and moderate-to-vigorous physical activity (MVPA). Objective functions include BMIz, cognitive performance (COG), life satisfaction (LS), and fitness (VO2). All objectives are modeled as maximization problems, except BMIz, which is formulated as minimizing its absolute value. These objective functions follow the same definitions as in~\cite{DBLP:conf/ppsn/XieNSRDN22}, and the corresponding regression
coefficients for the expected health benefit can be found in that work. 
The valid \textit{ilr} basis matrix used for the $4D$ problem is
$$
V^T =
\scalebox{0.95}{$
\begin{bmatrix}
-0.7071068 & 0.7071068 & 0.0000000 & 0.0000000\\
-0.4082483 & -0.4082483 & 0.8164966 & 0.0000000\\
-0.2886751 & -0.2886751 & -0.2886751 & 0.8660254
\end{bmatrix}$}
$$
\vspace{-2mm}
\begin{table}[h]
\begin{center}
\caption{The $(X^T X)^{-1}$ matrix used in the uncertainty function 
$U(\boldsymbol{x})=$
$U^*(\boldsymbol{x}_0)= s_e \sqrt{1 + \boldsymbol{x}_0^T (X^T X)^{-1} \boldsymbol{x}_0 }$ in the 4D problem for the Life Satisfaction outcome. Note that the $s_e$ value for the Life Satisfaction regression was 4634.088. 
}
\label{tab:lifeunc}
\vspace{2mm}
\scalebox{0.99}{
\begin{tabular}{l|rrrr}
\hline
  & (Intercept) & $z_1$ & $z_2$ & $z_3$\\
\hline
(Intercept) & 0.0144878 & -0.0139283 & 0.0002623 & -0.0001188\\
$z_1$  & -0.0139283 & 0.0410740 & -0.0142235 & -0.0058197\\
$z_2$  & 0.0002623 & -0.0142235 & 0.0142227 & -0.0046780\\
$z_3$  & -0.0001188 & -0.0058197 & -0.0046780 & 0.0105197\\
\hline
\end{tabular}}
\end{center}
\vspace{-6mm}
\end{table}

As an example, we provide the matrix used for the uncertainty quantification $U^*(x_0)$ of the health objective Life Satisfaction in Table~\ref{tab:lifeunc}.
Similar tables have been constructed for the other health objectives studied in our experimental investigations for the $4D$ as well as the $7D$ model.

\subsection{Experimental results for uncertainty quantification}
We evaluate uncertainty quantification by comparing the maps produced by the standard and discounted approaches for $10$ independent runs. Our algorithm is terminated after $10^6$ fitness evaluations. The $4D$ model does not allow for very different solutions for the different behavioral spaces. We will use the $4D$ model mainly to illustrate the ability of our approach to highlight the uncertainty associated with optimized solutions. To do this, we consider the standard set up using no discounting and display the expected health benefit as well as the uncertainty associated with it.
\subsubsection{The variable-based behavioral space}
\begin{figure}[t]
   \begin{subfigure}[t]{0.24\textwidth}
       \centering
       \includegraphics[width=.98\textwidth]{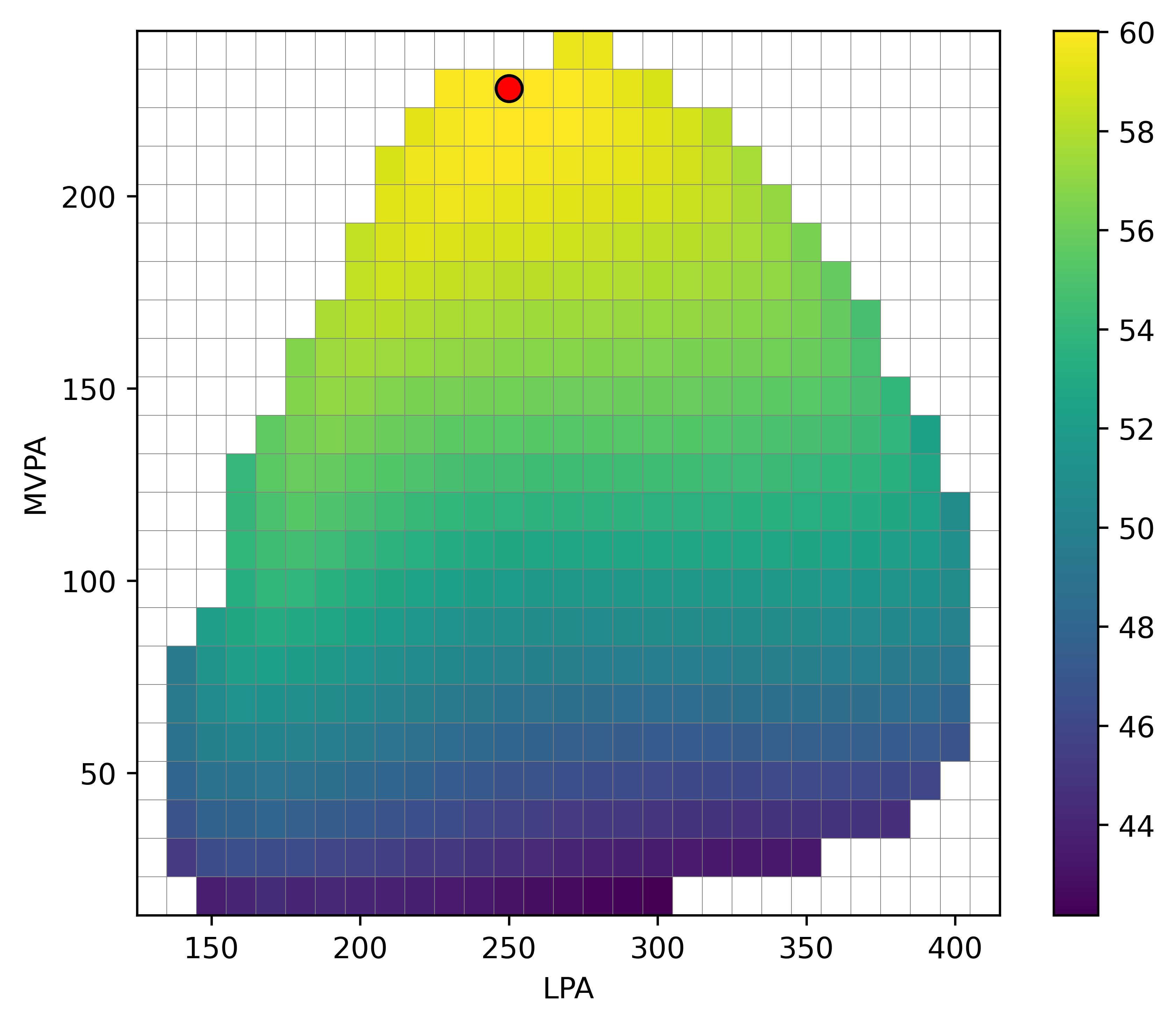}
        \caption{Obj=VO2, E(x)}
         \label{fig:VO-E-50}
         \end{subfigure}
        \begin{subfigure}[t]{0.24\textwidth}
        \centering
    \includegraphics[width=.98\textwidth]{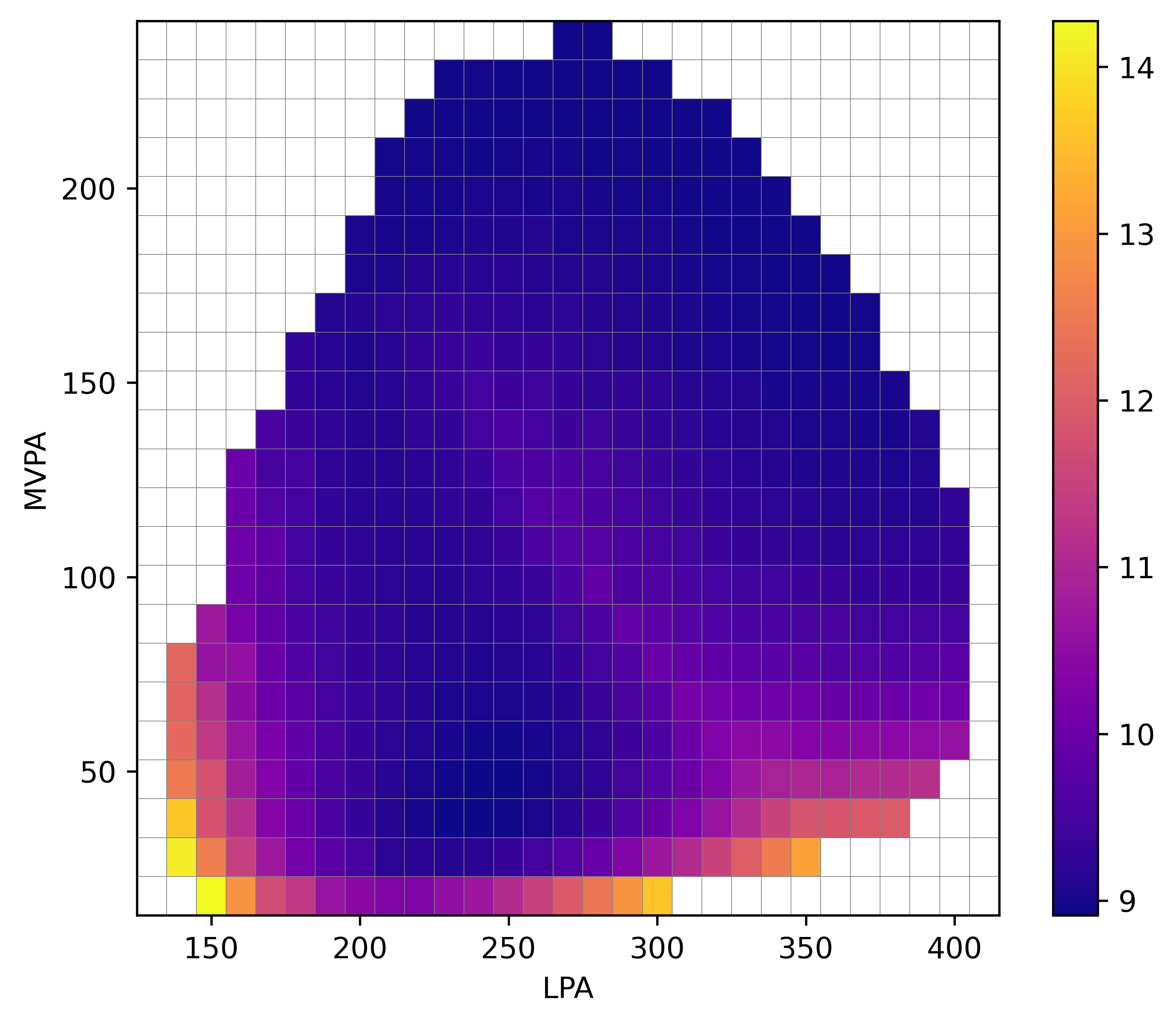}
    \caption{Obj=VO2, U(x)}
    \label{fig:VO-U-50}
     \end{subfigure}
        \begin{subfigure}[t]{0.25\textwidth}
        \centering
    \includegraphics[width=.98\textwidth]{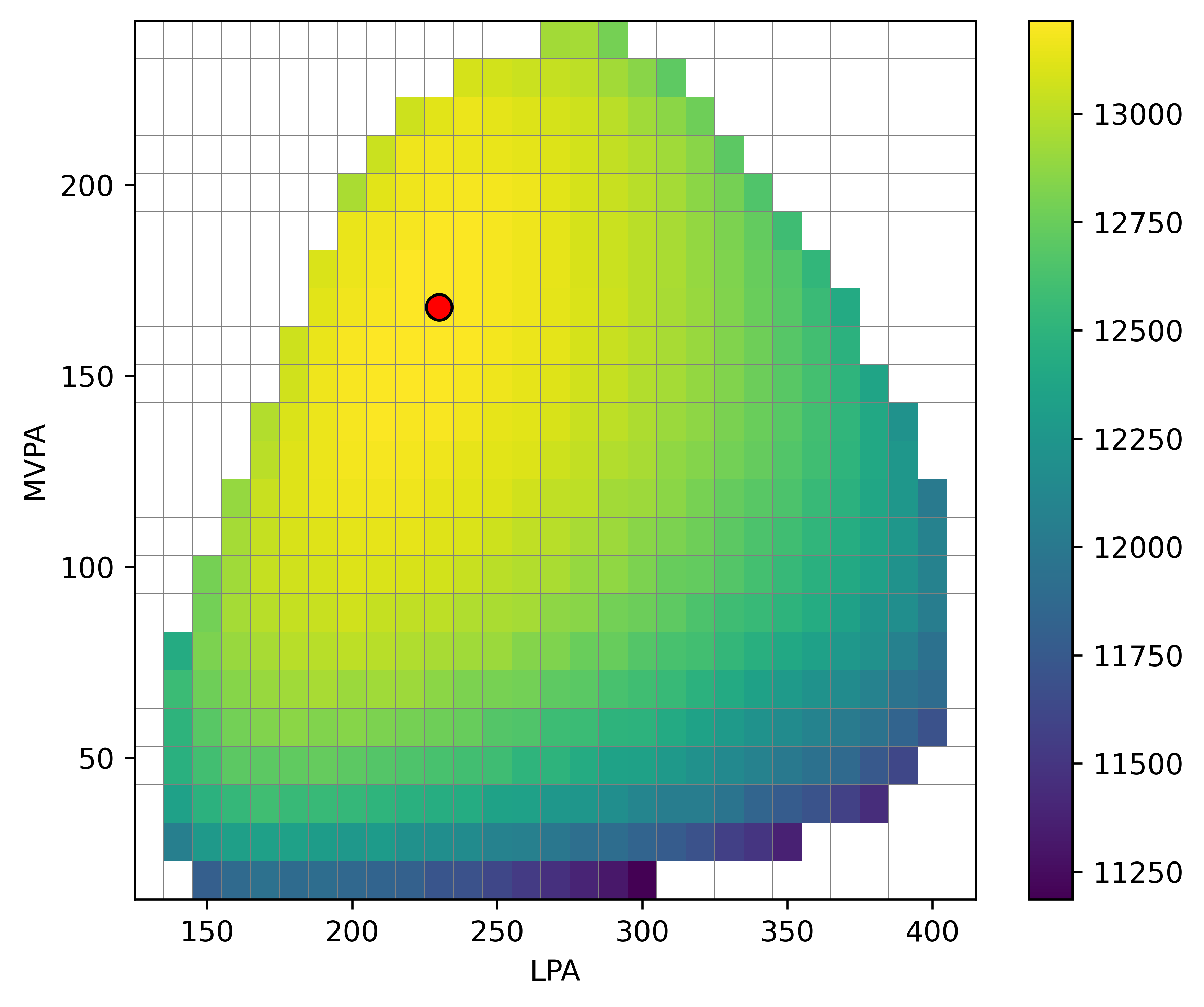}
         \caption{Obj=LS, E(x)}
             \label{fig:LS-E-50}
         \end{subfigure}
            \begin{subfigure}[t]{0.25\textwidth}
        \centering
   \includegraphics[width=.98\textwidth]{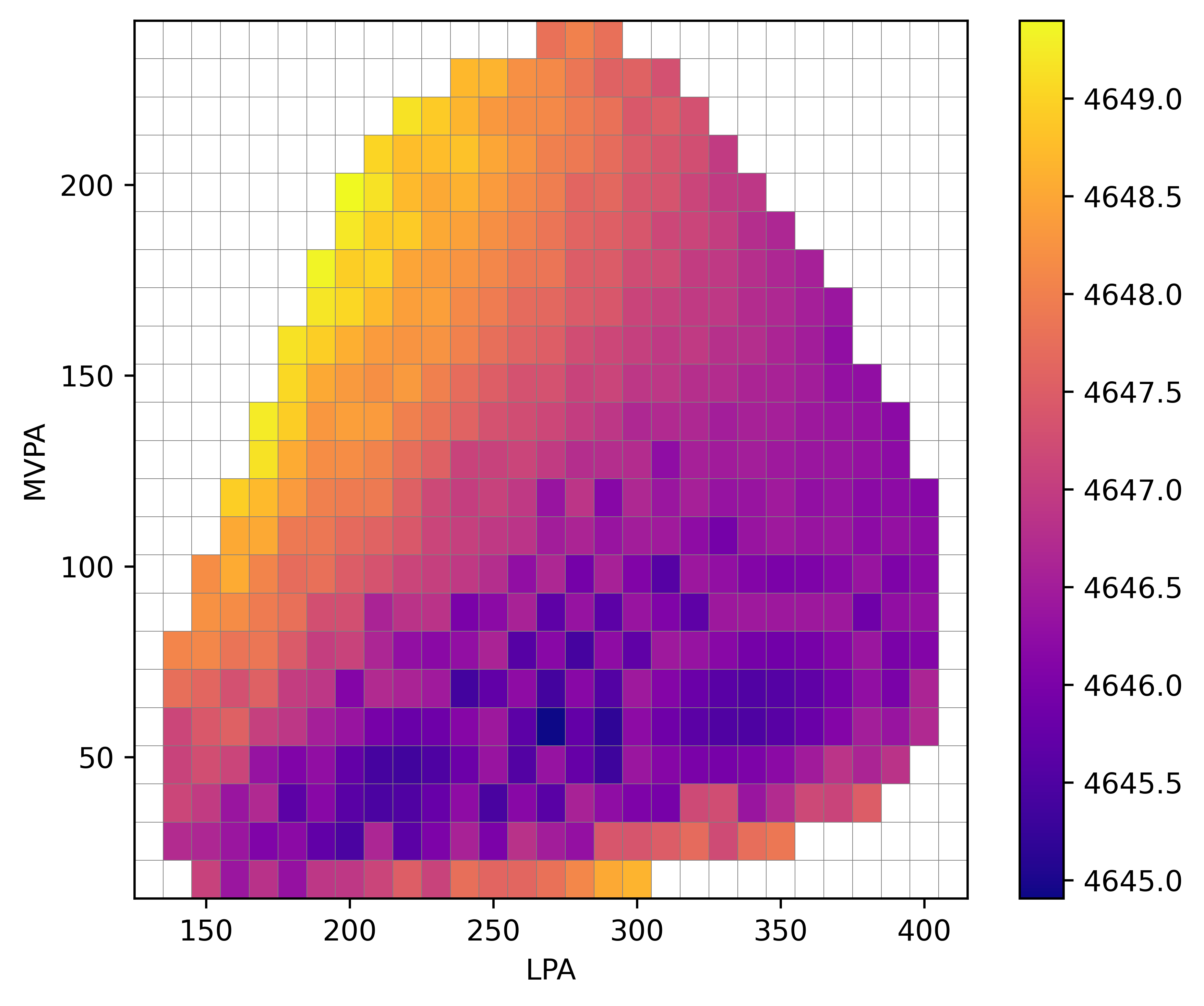}
          \caption{Obj=LS, U(x)}
         \label{fig:LS-U-50}
    \end{subfigure}\\
\begin{subfigure}[t]{0.24\textwidth}
        \centering
 \includegraphics[width=.98\textwidth]{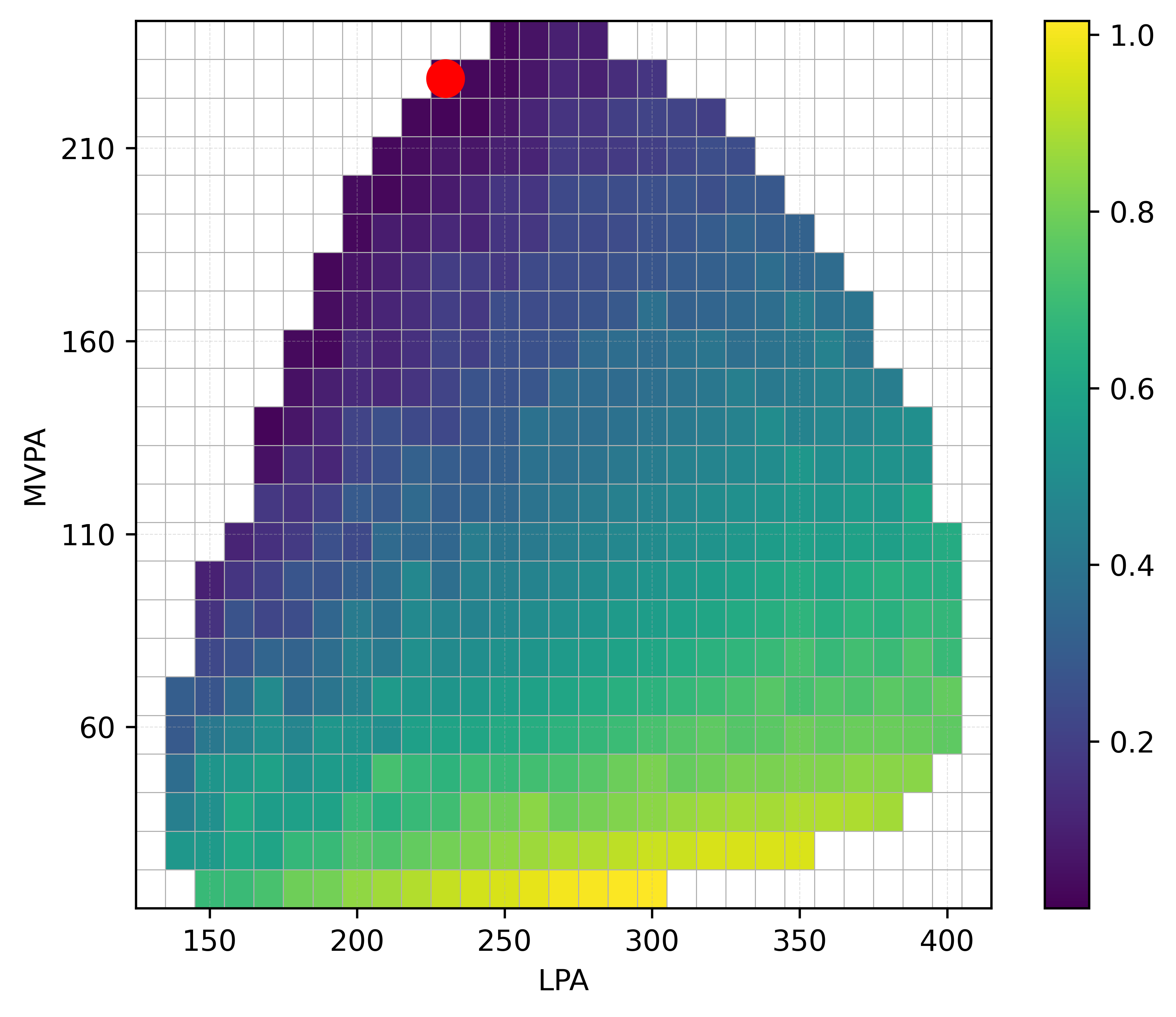} 
           \caption{Obj=BMIz, E(x)}
         \label{fig:BMIz-E-50}
    \end{subfigure}
\begin{subfigure}[t]{0.25\textwidth}
        \centering
 \includegraphics[width=.98\textwidth]{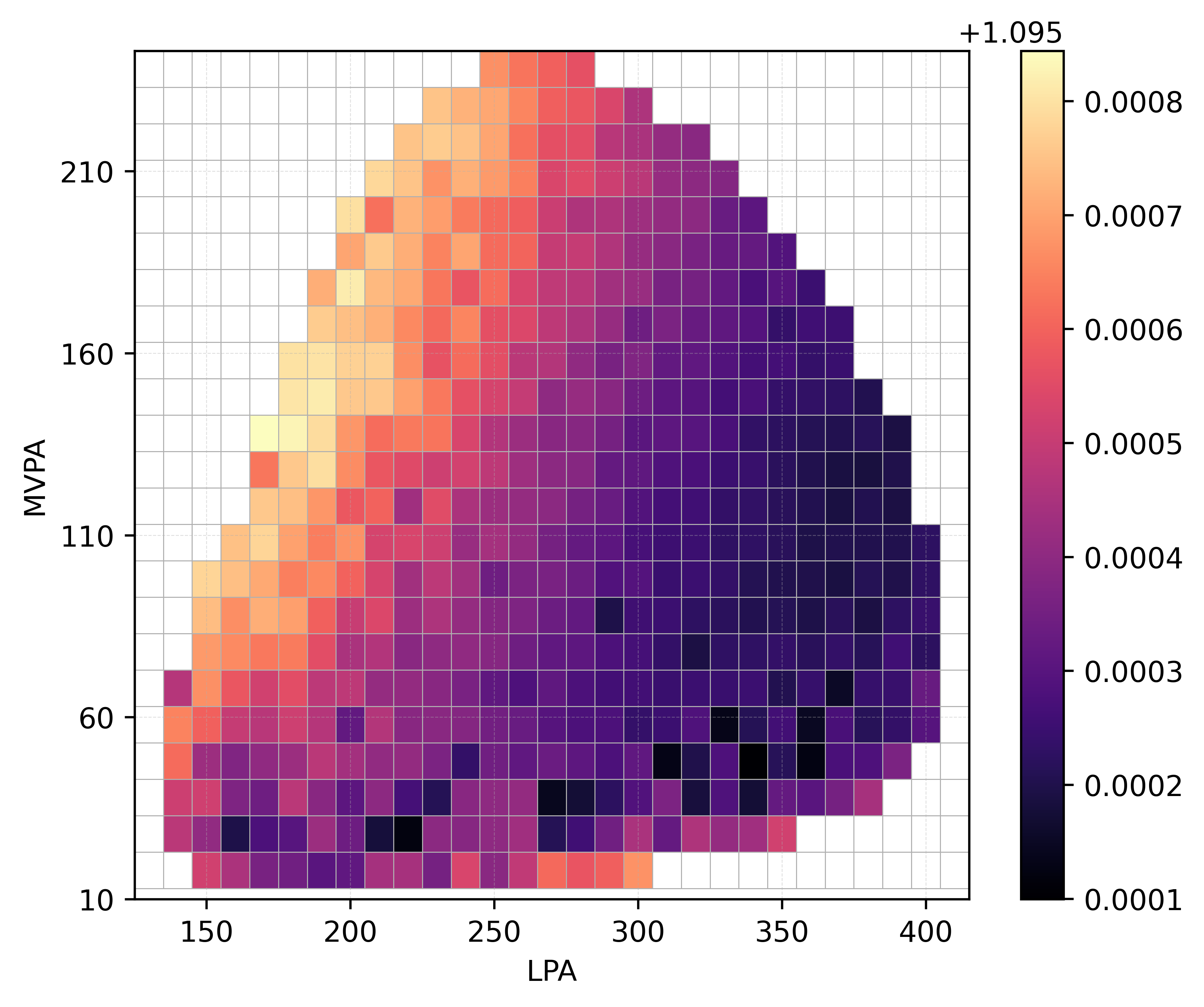}
          \caption{Obj=BMIz, U(x)}
         \label{fig:BMIz-U-50}
    \end{subfigure} 
    \begin{subfigure}[t]{0.238\textwidth}
        \centering
 \includegraphics[width=.98\textwidth]{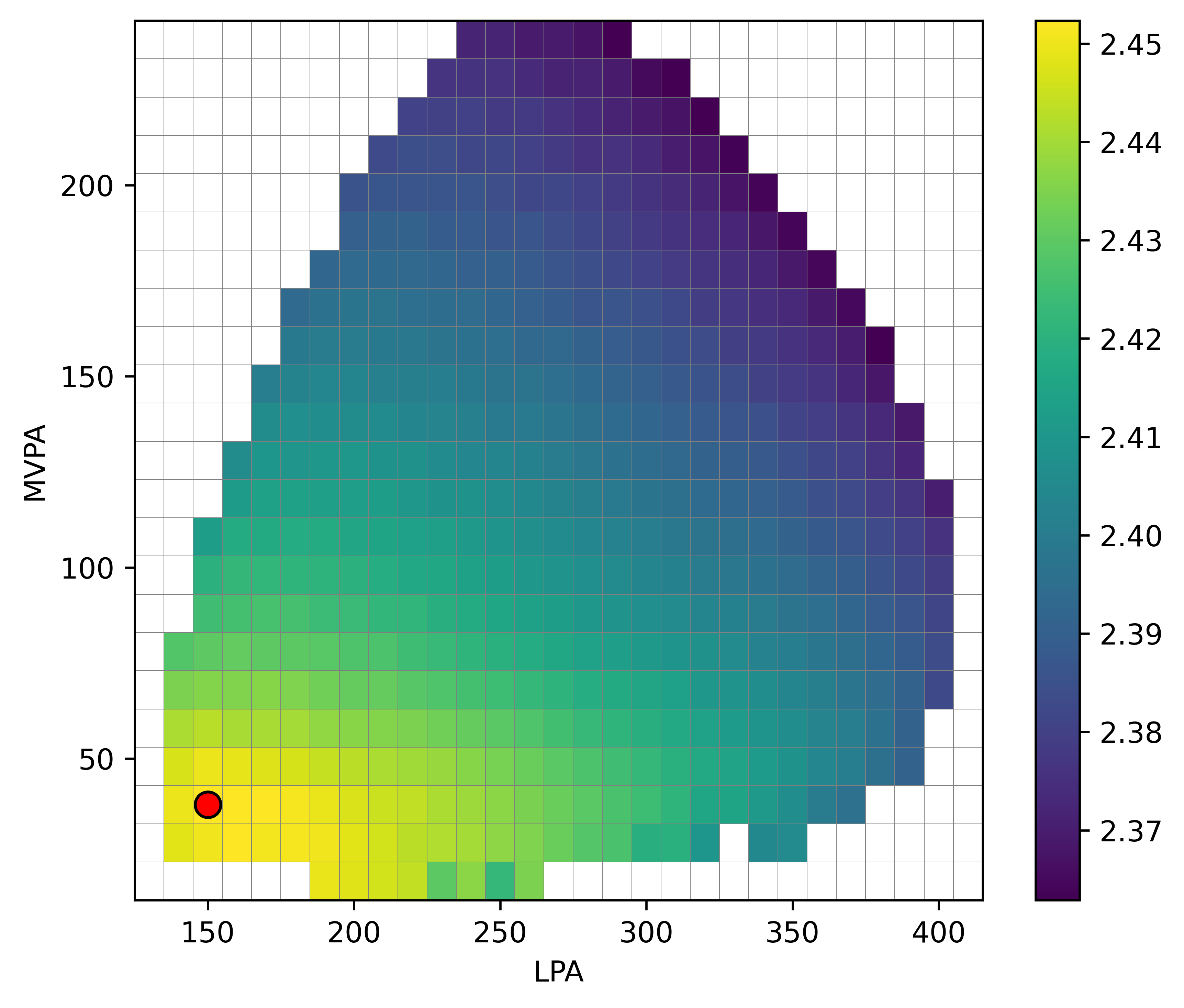} 
           \caption{Obj=COG, E(x)}
         \label{fig:Cog-E-50}
    \end{subfigure}
\begin{subfigure}[t]{0.25\textwidth}
        \centering
 \includegraphics[width=.98\textwidth]{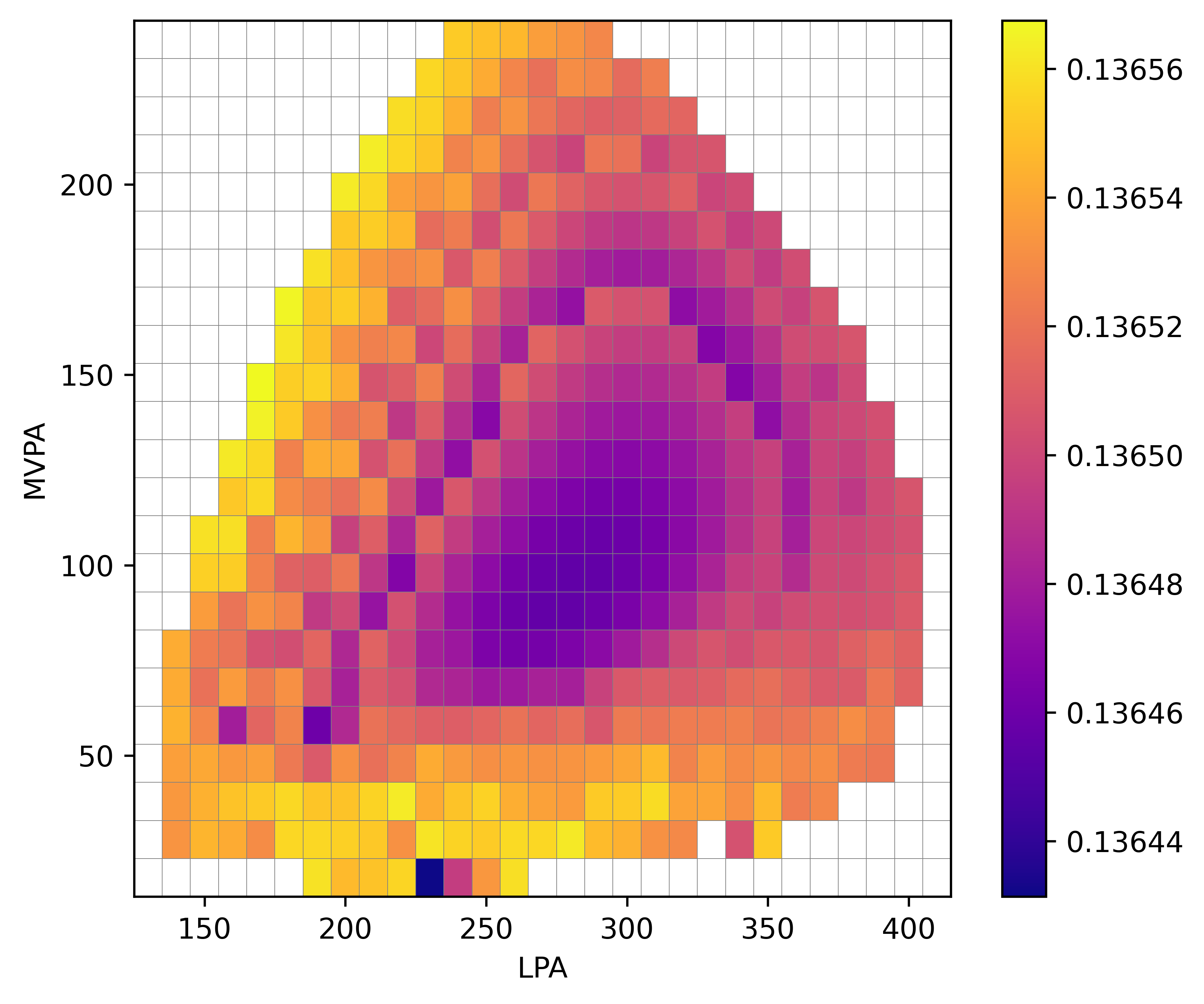}
          \caption{Obj=COG, U(x)}
         \label{fig:Cog-U-50}
    \end{subfigure}\\
\caption{
Distribution of expected health benefits and associated uncertainties for the 4D model in the VBS space (LPA, MVPA) using the standard approach.
}
\label{fig:Maps4DVBS}
\label{fig:VBS_4D}
\end{figure}
We show the results for the standard QD approach using the variable-based behavioral space (VBS) given by LPA and MVPA for the different health objectives in Figure~\ref{fig:Maps4DVBS}. For the different objectives, we can observe the expected health benefit as well as the uncertainty associated with the corresponding solution.
The best solution for the standard approach (red dot in Figure~\ref{fig:VO-E-50}) has a VO2 value of $E(x)$=$60.0145$ and $U(x)$=$8.9677$, corresponding to the time-use elite [$514$,$450$,$261$,$215$]. 
In Figures~\ref{fig:LS-E-50} and~\ref{fig:LS-U-50}, the approach achieves an LS value of $E(x)$=$13215.0981$ with $U(x)$=$4648.4160$ for the elite solution [$613$,$436$,$230$,$161$]. 
In Figures~\ref{fig:BMIz-E-50} and~\ref{fig:BMIz-U-50}, the method identifies the elite [$557$,$536$,$197$,$150$] with a BMIz value of $E(x)$=$0.00001$ and uncertainty $U(x)$=$1.0958$. 
The approach achieves a COG value $E(x)$=$2.4523$ with $U(x)$=$0.1365$ at the solution [$527$,$738$,$159$,$16$], as shown in Figures~\ref{fig:Cog-E-50} and~\ref{fig:Cog-U-50}.
Under the discounted approach, the resulting maps remain similar to those obtained under the standard approach. This is primarily due to the small number of behavioral variables, which strongly constrains the structure of the solution space. As a result, even under strong discounting, the overall patterns and identified high-performing regions stay largely unchanged, with only marginal shifts observed in a few local areas.

Our results show that the relationship between expected performance and uncertainty is strongly objective-specific. 
For example, for VO2, high-quality solutions are consistently associated with low uncertainty. This indicates a stable region of the learned model where optimal solutions are robust to uncertainty. 
In contrast, LS and BMIz exhibit a different uncertainty-quality patterns, where high-quality solutions are linked to higher uncertainty. This suggests that optimal regions for these objectives arise partly in areas of weaker model support or greater extrapolation. Uncertainty discounting slightly changes the selected best solutions and does not remove this pattern, indicating that performance and uncertainty are closely connected. COG exhibits a third behavior in which the objective landscape shows little variation in performance across solutions. In this case, uncertainty has limited influence on ranking and mainly serves as a secondary criterion for distinguishing between near-equivalent solutions.
Overall, the results reveal three distinct regimes: low-uncertainty reliable optima (VO2), uncertainty-driven high-performance regions (LS, BMIz), and low-discriminative objectives (COG). This shows that uncertainty in QD serves as an indicator of model reliability rather than a uniform penalty, and it reshapes the archive differently for each objective while still preserving diversity.
\begin{figure*}[t!]
   \begin{subfigure}[t]{0.243\textwidth}
       \centering
        \includegraphics[width=.99\textwidth]{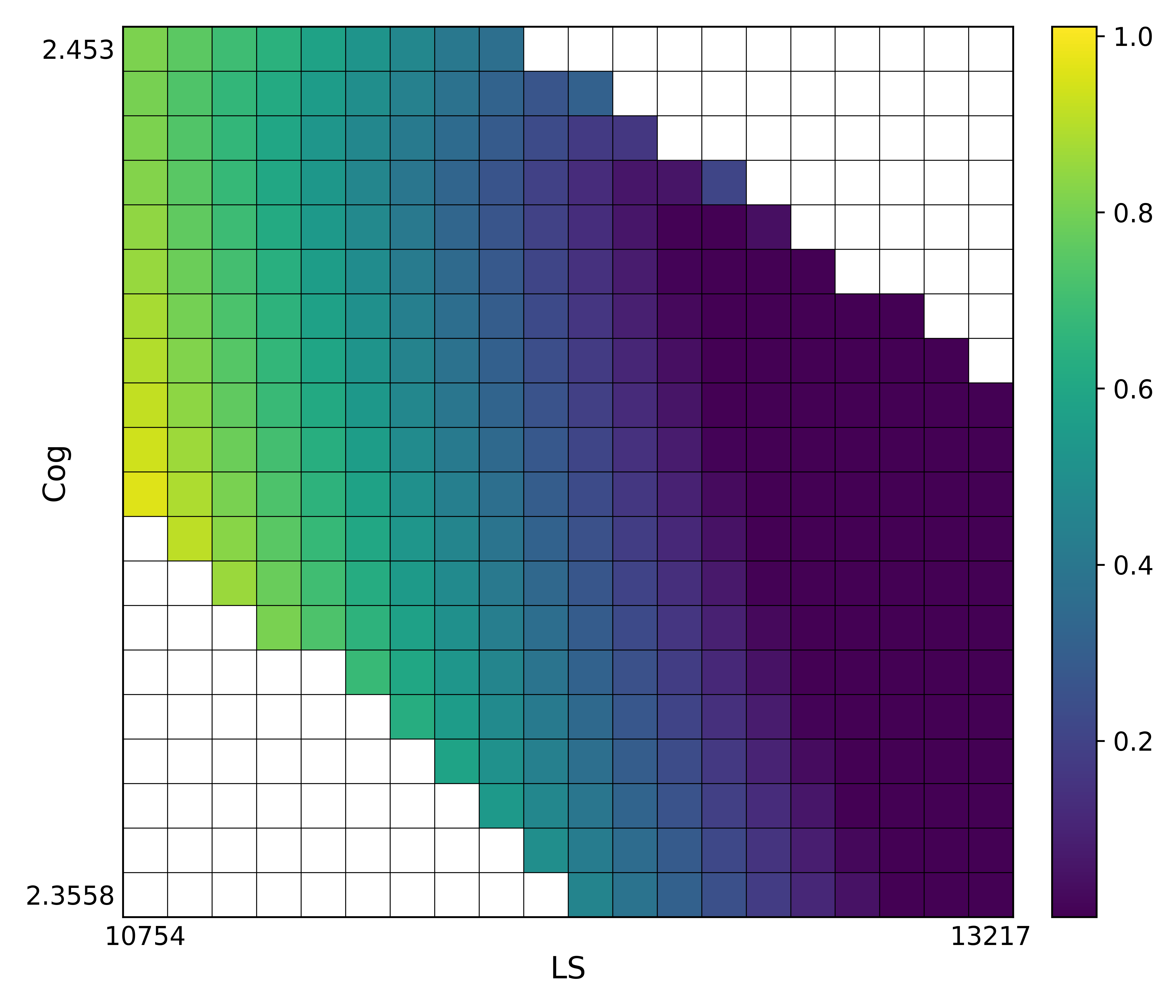}
        \caption{Obj=BMIz, E(x)}
         \end{subfigure}
        \begin{subfigure}[t]{0.243\textwidth}
        \centering
    \includegraphics[width=.99\textwidth]{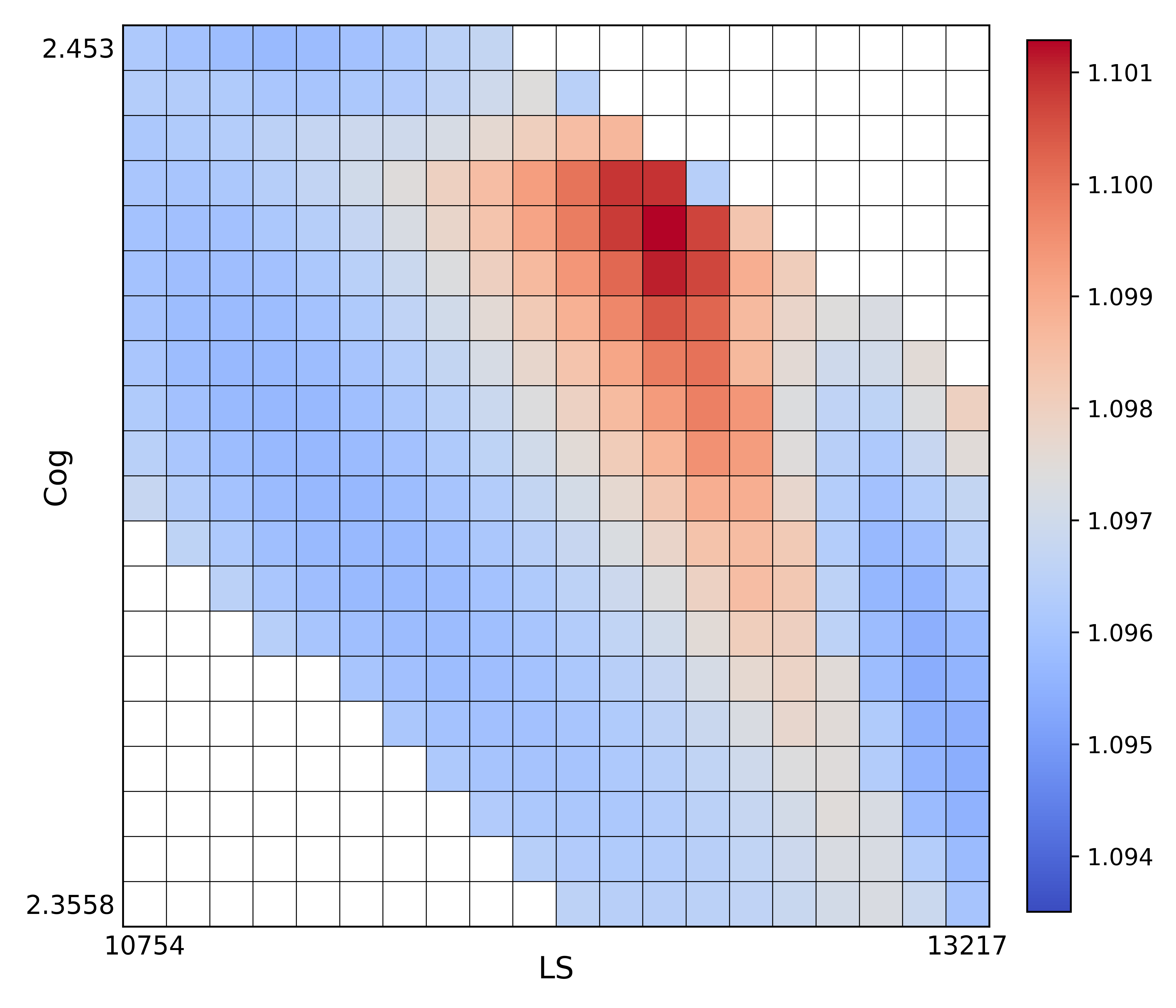}
    \caption{Obj=BMIz, U(x)}
     \end{subfigure}
        \begin{subfigure}[t]{0.243\textwidth}
        \centering
        \includegraphics[width=.99\textwidth]{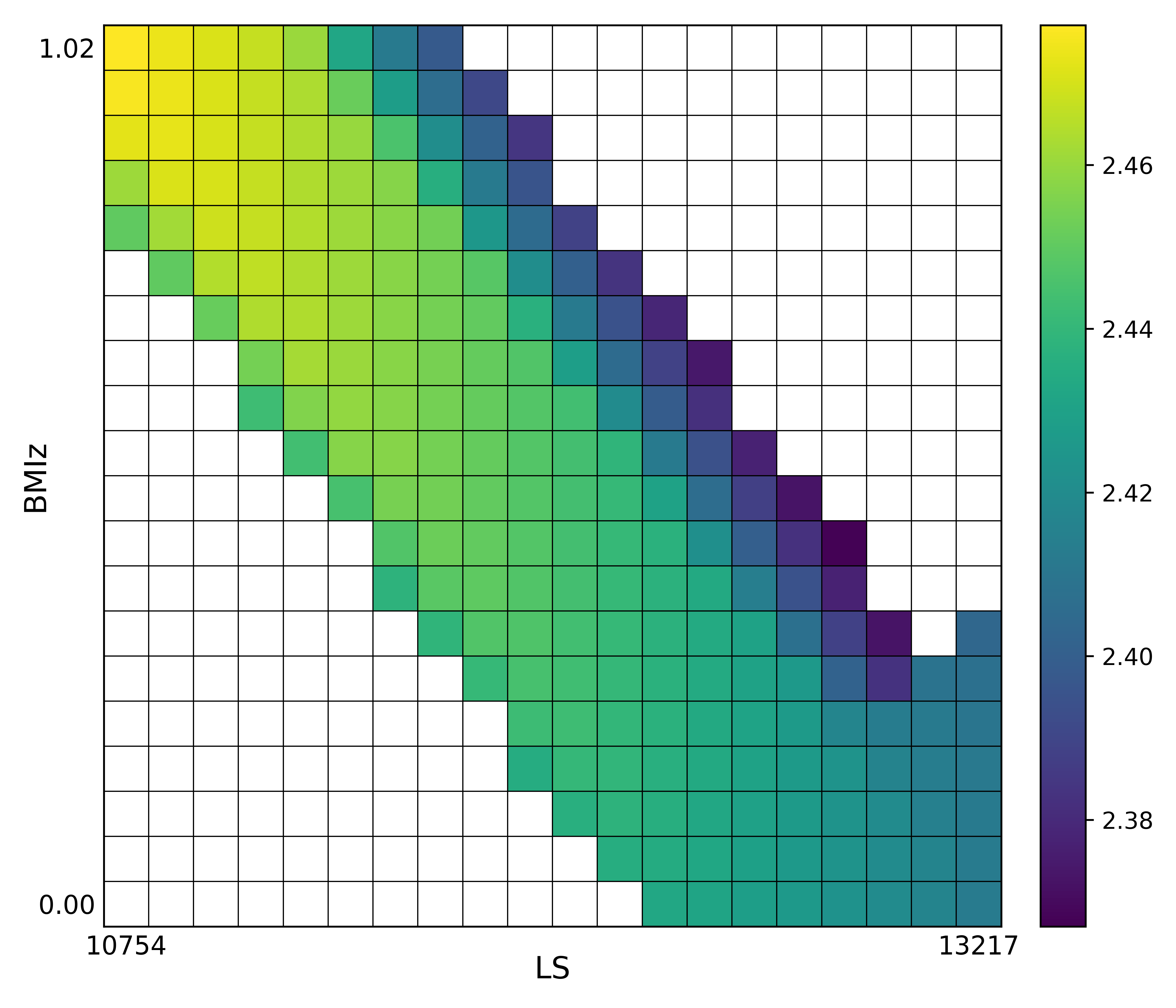}
         \caption{Obj=COG, E(x)}
           \label{fig:COG_E_50}
         \end{subfigure}
            \begin{subfigure}[t]{0.243\textwidth}
        \centering
   \includegraphics[width=.99\textwidth]{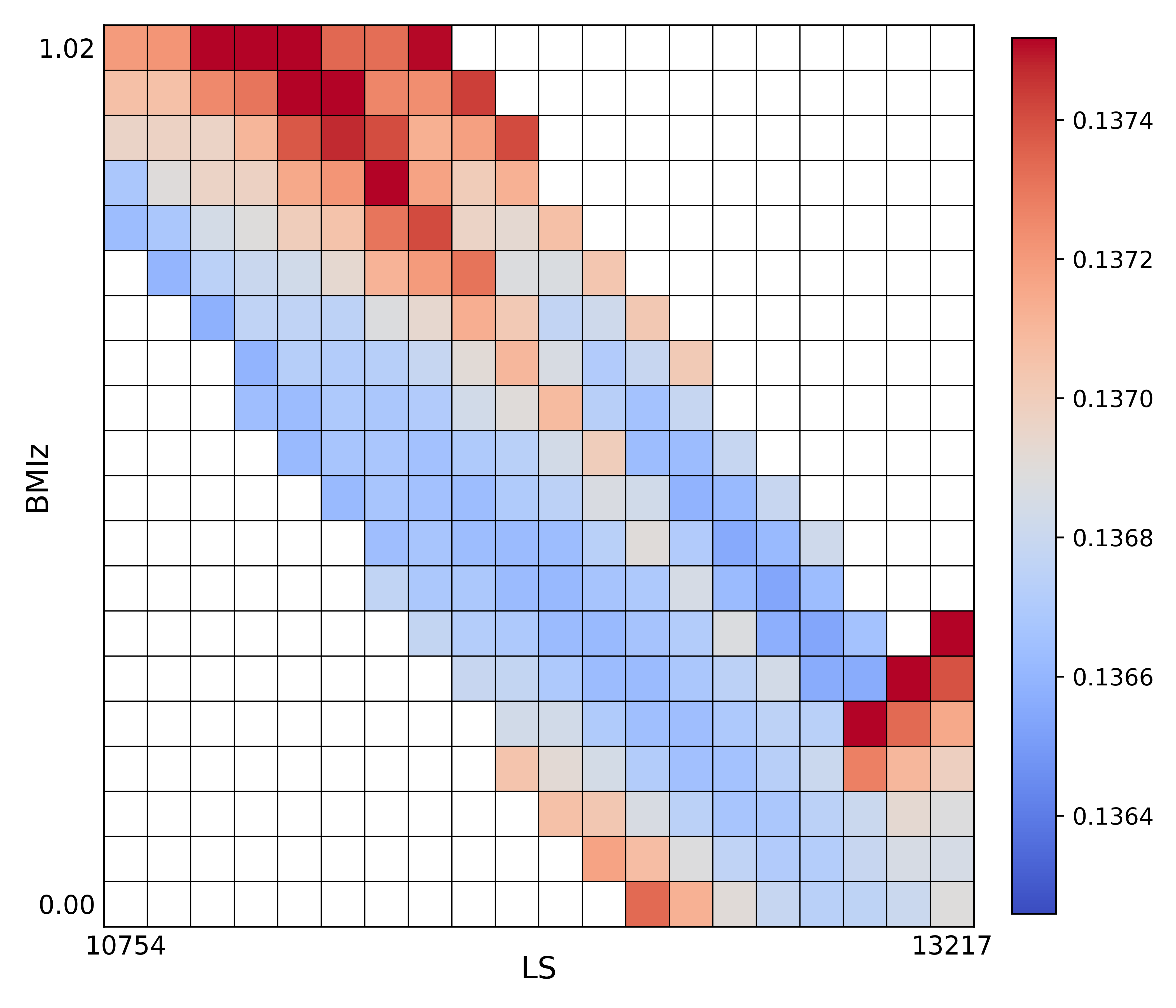}
          \caption{Obj=COG, U(x)}
             \label{fig:COG_U_50}
    \end{subfigure}
\caption{
Distribution of expected health benefits and associated uncertainties for the 4D model in the OBS space (LS, Cog) (left two columns and (LS, BMIz) (right two columns) using the standard approach.
}
\label{fig:OBS_4D}
%\vspace{-4mm}
\end{figure*}

\subsubsection{The objective-based behavioral space}
We evaluate the performance of our algorithm within the objective-based behavioral space (OBS). Figure~\ref{fig:OBS_4D} illustrates the elite solutions identified by the standard QD approach within two objective-based behavioral spaces such as COG x LS, BMIz x LS, where the objective functions correspond to Body Mass Index (BMI z-score) and cognitive performance (COG), respectively. 
Instead of giving one recommendation, our approach provides a map of elite behavioral compositions, together with their associated uncertainty.
This allows for personalized health interventions, for example if a person cannot increase MVPA, the map shows the next best elite composition within their feasible sedentary or LPA range, while also indicating the reliability of the predicted outcome. OBS map is useful for finding the most certain behaviors by mapping one health objective against another, systematically identifying regions with high performance and low uncertainty across the space.
Figure~\ref{fig:OBS_4D} shows that regions with higher cognitive scores and higher life satisfaction correspond to higher BMIz values. The best-performing solution in this space has $E(x)$=$0.9626$ and $U(x)$=$1.0968$ at position ($20$,$10$). It also highlights that there are no solutions where both cognitive and life satisfaction values are simultaneously very low or very high.
Furthermore, Figure~\ref{fig:OBS_4D} shows a clear contrast in the uncertainty patterns for BMIz and COG. On the another hand, the discounted strategy leads to a similar uncertainty patterns, while the expected value maps for BMIz and COG remain largely unchanged. The discounted approach reduces uncertainty in specific regions of the space, indicating improved robustness without substantially altering the overall structure of the expected value landscape.
Figures~\ref{fig:COG_E_50} and~\ref{fig:COG_U_50} show the $E(x)$ and $U(x)$ maps obtained for cognitive function as the objective. The feasible region in this space is more constrained than in the BMIz space, indicating stronger behavioral restrictions. The results suggest that higher cognitive performance is associated with increased BMIz and reduced LS. The best cognitive value achieved is $E(x)$=$2.4771$ and $U(x)$=$0.1372$ at position ($1$,$20$).

\subsection{The 7D problem}

We analyze the performance of the QD algorithms on the $7D$ problem, where the behavioral space consists of sleep, screen time, physical activity, quiet time, passive transport, school-related activities, and domestic/self-care time. The optimization involves three objective functions such as academic performance, body fat percentage and psychiatric score~\cite{dumuid2022yourbestday}.
We use this setting to further investigate how uncertainty propagates across the high-dimensional behavioral space and interacts with solutions.
In particular, we analyze how our discounted approach can effectively reduce uncertainty, typically at the cost of lower expected health benefit. We further compare the standard and discounted approaches in terms of expected value and uncertainty in order to enable more informed and practical decision-making for behavioral health intervention.

\begin{figure}[t]
   \vspace{1cm}
   \begin{subfigure}[t]{0.325\textwidth}
       \centering
 \includegraphics[width=0.99\textwidth]
 {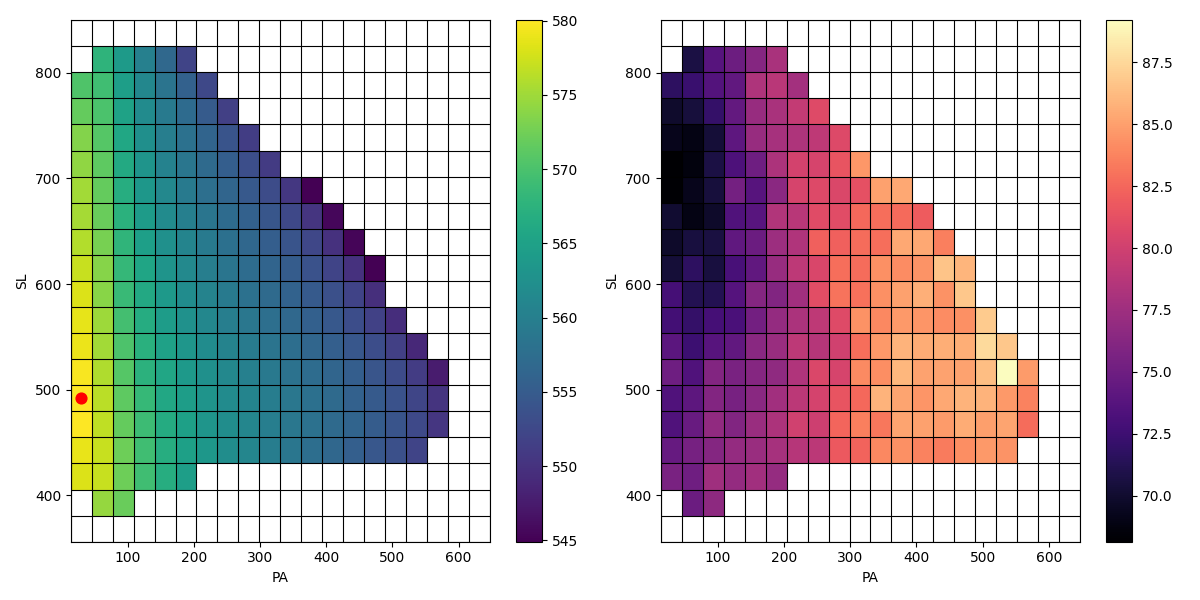}
        \caption{Obj=Acad, E(x), U(x)}
         \label{fig:Acad_50}
         \end{subfigure}
 \begin{subfigure}[t]{0.325\textwidth}
       \centering  
        \includegraphics[width=0.99\textwidth]
        {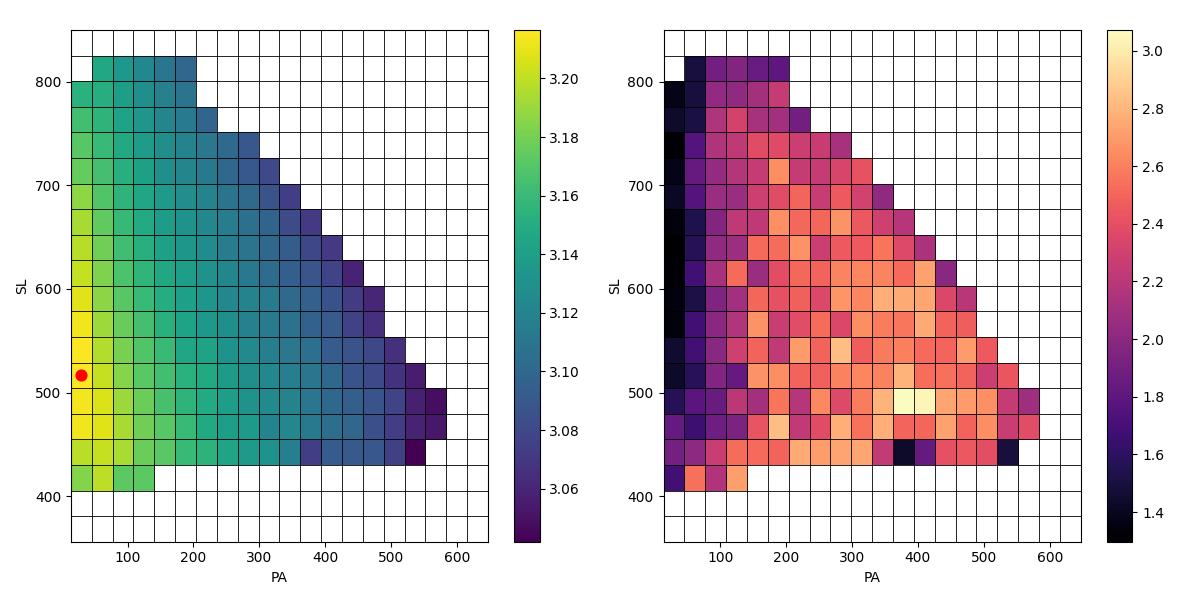}
             \caption{Obj=BF, E(x), U(x)}
         \label{fig:BF_50}
         \end{subfigure}
   \begin{subfigure}[t]{0.325\textwidth}
       \centering      
     \includegraphics[width=0.99\textwidth]
     {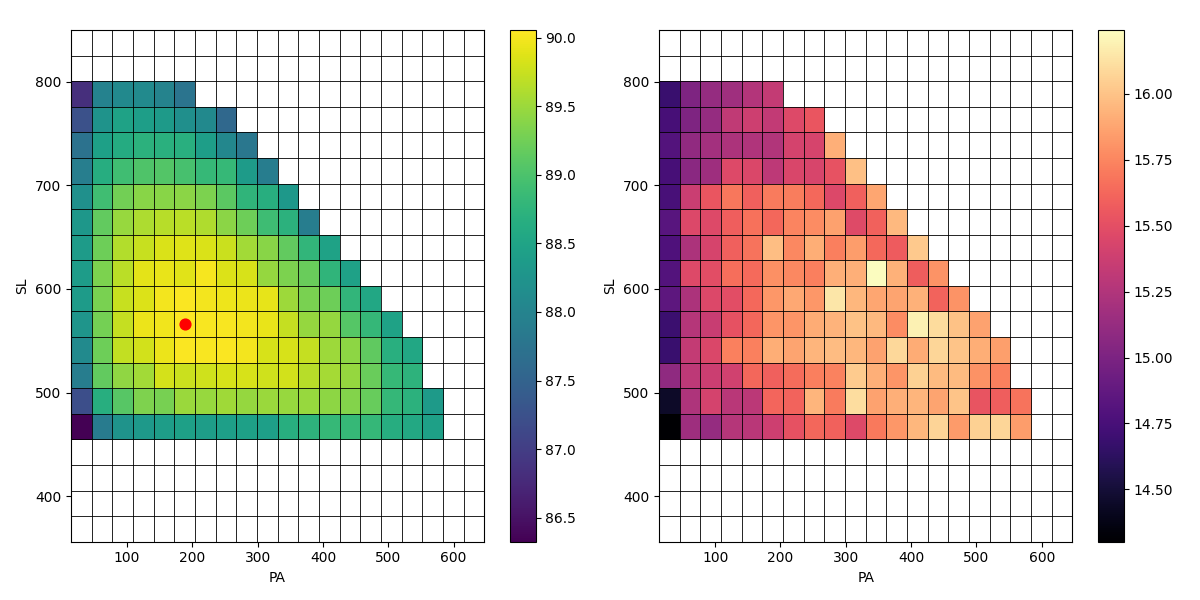}
         \caption{Obj=Psy, E(x), U(x)}
         \label{fig:Psy_50}
         \end{subfigure}\\ 
   \begin{subfigure}[t]{0.325\textwidth}
       \centering
    \includegraphics[width=0.99\textwidth] 
    {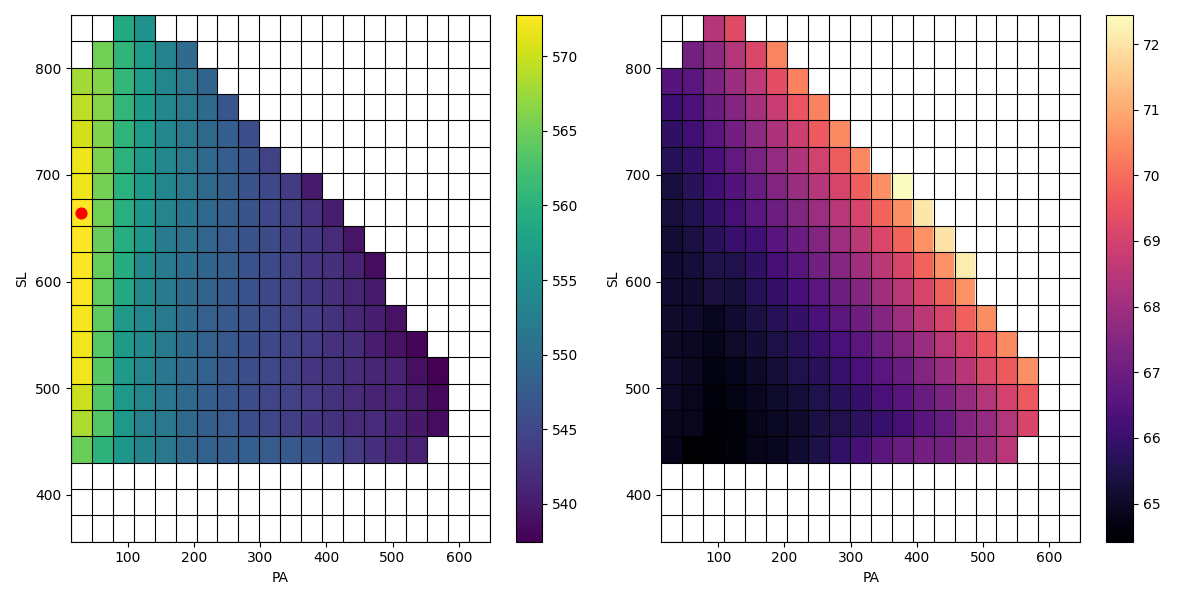} 
        \caption{Obj=Acad, E(x), U(x)}
         \label{fig:Acad_10_d}
         \end{subfigure}
       \begin{subfigure}[t]{0.325\textwidth}
       \centering
        \includegraphics[width=0.99\textwidth]
        {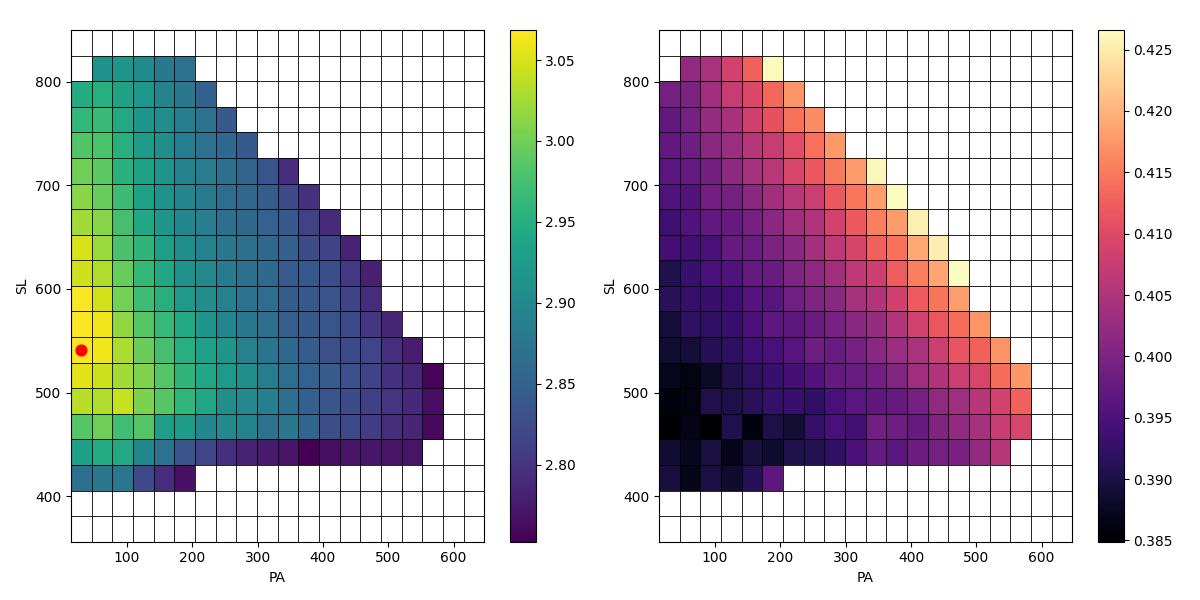}
                     \caption{Obj=BF, E(x), U(x)
                     }
         \label{fig:BF_10_d}
         \end{subfigure}
   \begin{subfigure}[t]{0.325\textwidth}
       \centering
       \includegraphics[width=0.99\textwidth]
       {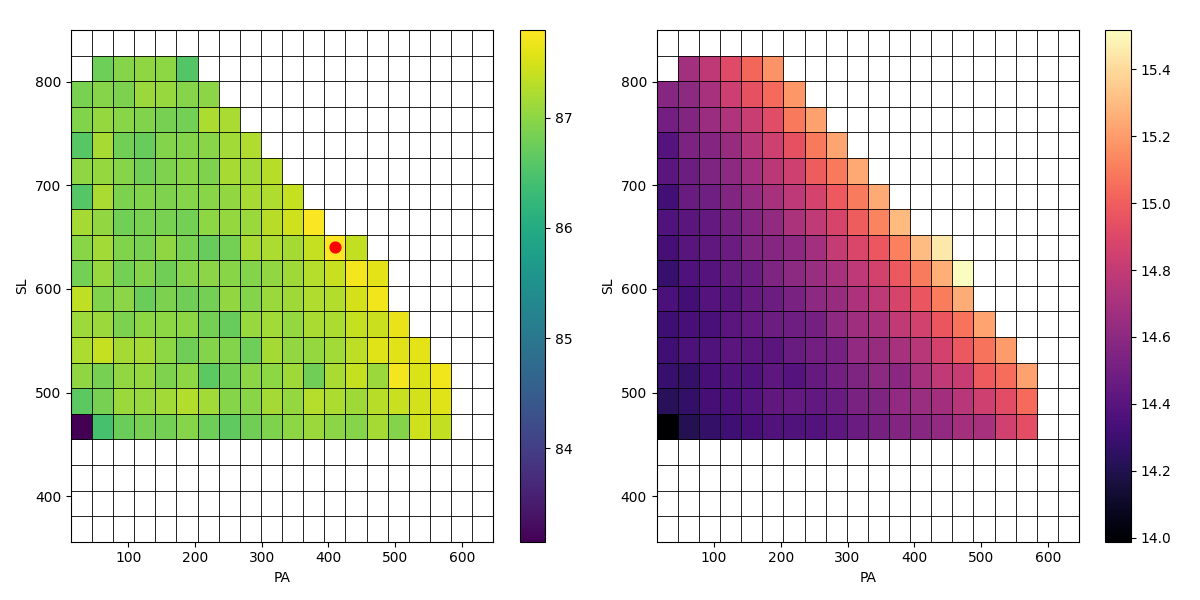}
                \caption{Obj=Psy, E(x), U(x)}
         \label{fig:Psy_10_d}
         \end{subfigure}
 \caption{Distribution of expected health benefits and associated uncertainties for the 7D model in the VBS space (PA, SL) using the standard approach (top) and discounted approach with $C_{\alpha}=6.36134$ (bottom).
}
    \label{fig:VBS_7D} 
\end{figure}

\subsubsection{The variable-based behavioral space}
For the $7D$ optimization problem, the search space becomes substantially larger. To ensure convergence within a reasonable number of fitness evaluations, we restrict the analysis to $2D$ behavioral spaces. For each cell, we compute the expected value and uncertainty.
Figure~\ref{fig:VBS_7D} presents the maps obtained using the standard and discounted approach, with behavioral spaces defined by physical activity × sleep. We note that the discounting approach leads to a significant reduction in uncertainty and the color-coding for the two approaches is on different scales due to this.
Across all objectives, the maps show that high quality solutions are not evenly distributed but instead concentrate in specific regions of the space. The relationship between expected value $E(x)$ and uncertainty $U(x)$ is objective-specific. High-quality solutions are localized near the boundaries of the behavioral space for academics and body fat, whereas the interior is dominated by solutions with moderate performance and a wide range of uncertainty levels. Here, higher performance aligns with low uncertainty, and both approaches identify similar high-quality regions. 
For example, in Figure~\ref{fig:Acad_50} and~\ref{fig:BF_50}, the elite [$480$,$1$,$28$,$95$,$137$,$609$,$90$] for academic performance achieves $E(x)$=$580.03$, $U(x)$=$73.41$, while the elite [$510$,$281$,$24$,$284$,$1$,$320$,$20$] for body fat achieves $E(x)$=$3.22$ and $U(x)$=$1.53$, respectively. 
Figure~\ref{fig:Acad_10_d} and~\ref{fig:BF_10_d} presents the elite [$554$,$1$,$2$1,$13$,$190$,$418$,$243$] for academic performance with $E(x)$=$572.27$ and $U(x)$=$65.08$,
and the elite [$529$,$28$,$34$,$128$,$268$,$123$,$330$] for body fat with $E(x)$=$3.07$ and $U(x)$=$0.39$, respectively.
In contrast, for psychiatric, high-quality regions are concentrated in areas of elevated uncertainty. The elite solution [$569$,$1$,$242$,$123$,$403$,$26$,$76$] gives $E(x)$=$90.04$, $U(x)$=$15.80$ (see Fig.~\ref{fig:Psy_50}), and the elite [$520$,$1$,$35$,$17$,$616$,$70$,$181$] has $E(x)$=$87.03$, $U(x)$=14.29 (see Fig.~\ref{fig:Psy_10_d}). In this objective, high-quality solutions are predominantly located in regions of higher uncertainty.
Our discounted approach preserves the overall structure and location of high-quality regions but shifts solutions toward lower uncertainty. High-uncertainty extremes become less prominent, while the main performance patterns remain unchanged. This effect varies across objectives, with minimal changes in stable regions and more noticeable redistribution where performance is closely related to uncertainty.
For psychiatric, both approaches show noticeable differences. While similar performance regions are observed, the discounted approach shifts solutions toward the boundaries and reduces high uncertainty areas. In contrast, the standard approach shows a wider spread in uncertainty, with solutions more concentrated near boundary regions associated with lower physical activity.
In a nutshell, the results demonstrate that the discounted approach leads to a strong reduction in uncertainty across solutions in an objective-specific manner. It reveals distinct uncertainty performance relationships for different objectives and systematically shifts solutions toward lower uncertainty regions while preserving the overall structure of the search space.

\begin{figure}[t!]  
    \centering  
\begin{subfigure}[t]{0.24\textwidth}
        \centering
   \includegraphics[width=0.98\textwidth]{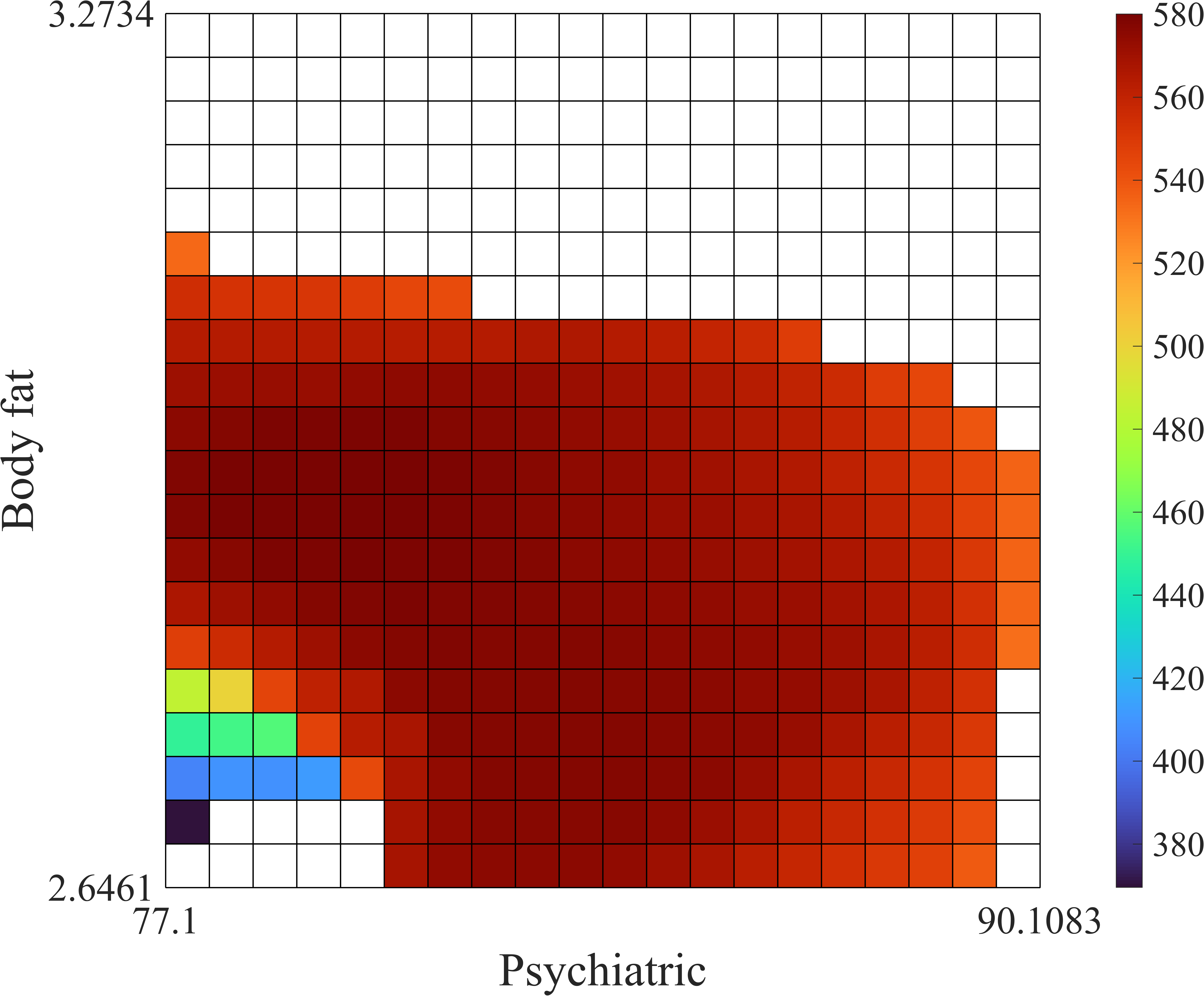}  
          \caption{Obj=Acad, E(x)}
         \label{fig:Acad-E-50}
    \end{subfigure}   
\begin{subfigure}[t]{0.24\textwidth}
        \centering
      \includegraphics[width=0.98\textwidth]{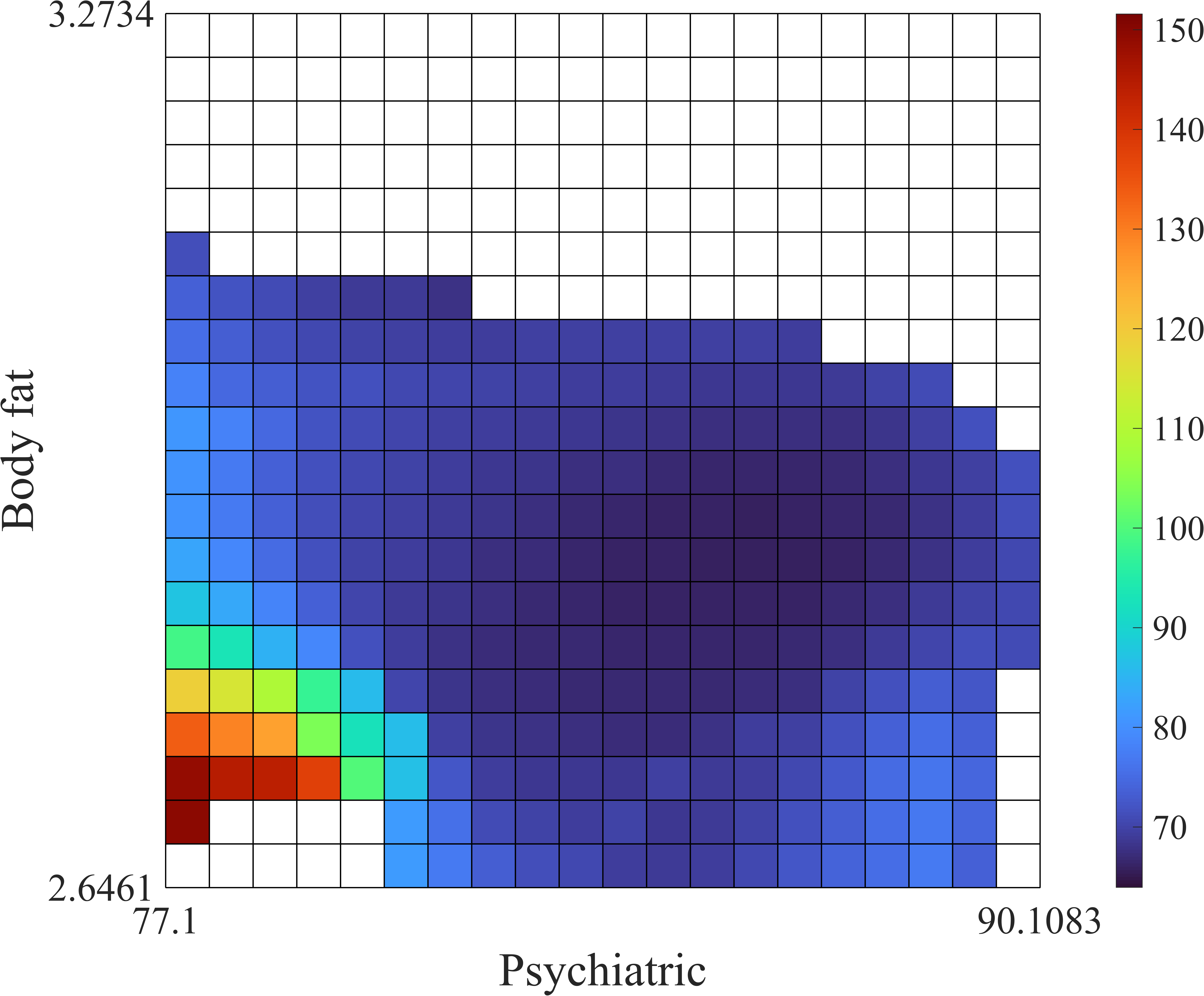}  
          \caption{Obj=Acad, U(x)}
         \label{fig:Acad-U-50}
    \end{subfigure}     
\begin{subfigure}[t]{0.24\textwidth}
        \centering
       \includegraphics[width=0.98\textwidth]{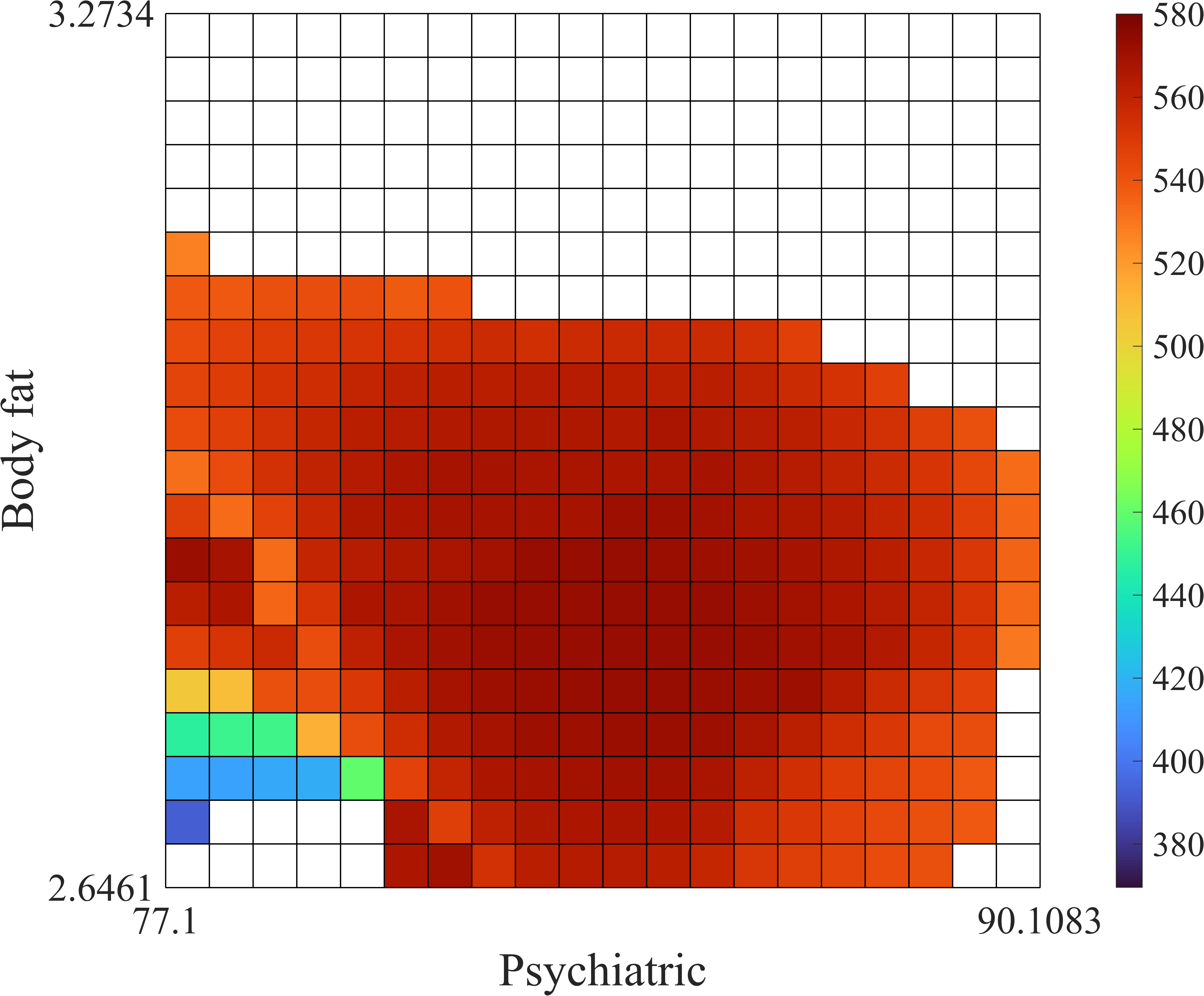}
          \caption{Obj=Acad, E(x)}
         \label{fig:Acad-E-10}
    \end{subfigure}
\begin{subfigure}[t]{0.24\textwidth}
        \centering
      \includegraphics[width=0.98
   \textwidth]{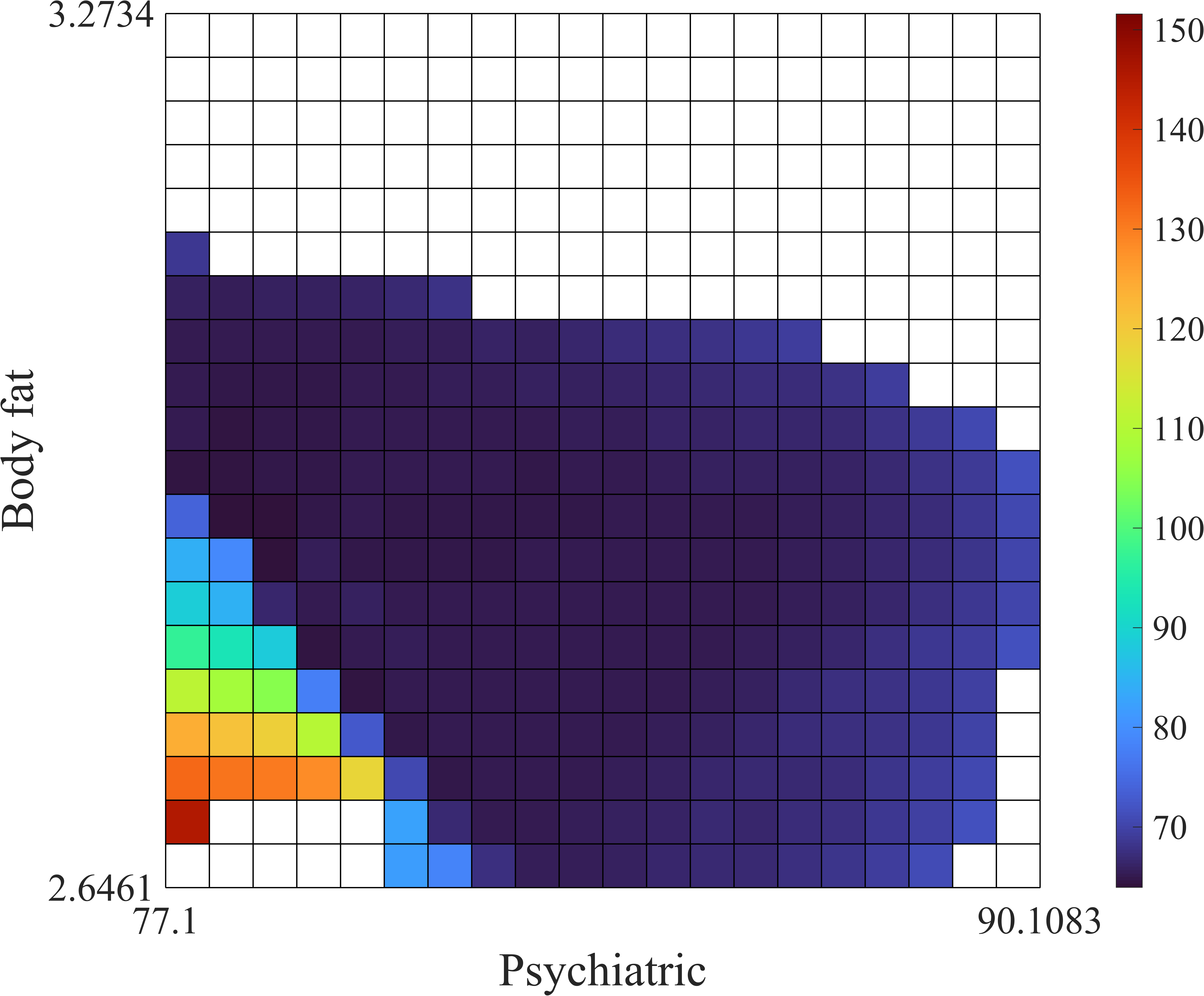}  
          \caption{Obj=Acad, U(x)}
         \label{fig:Acad-U-10}
    \end{subfigure}     
\begin{subfigure}[t]{0.24\textwidth}
        \centering
        \includegraphics[width=0.98\textwidth]{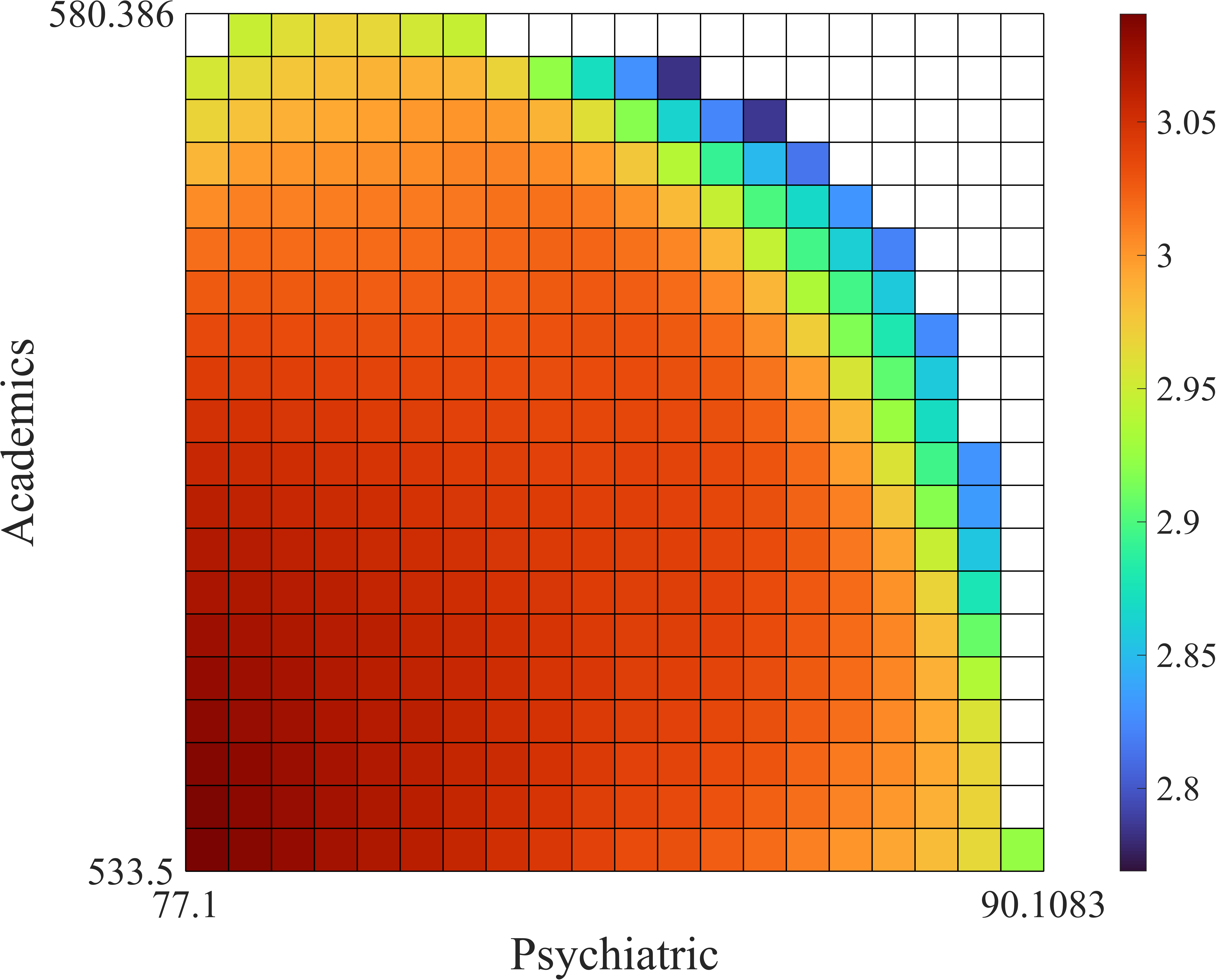}  
          \caption{Obj=BF, E(x)}
         \label{fig:BF-E-50}
    \end{subfigure}  
\begin{subfigure}[t]{0.24\textwidth}
        \centering
        \includegraphics[width=0.98\textwidth]{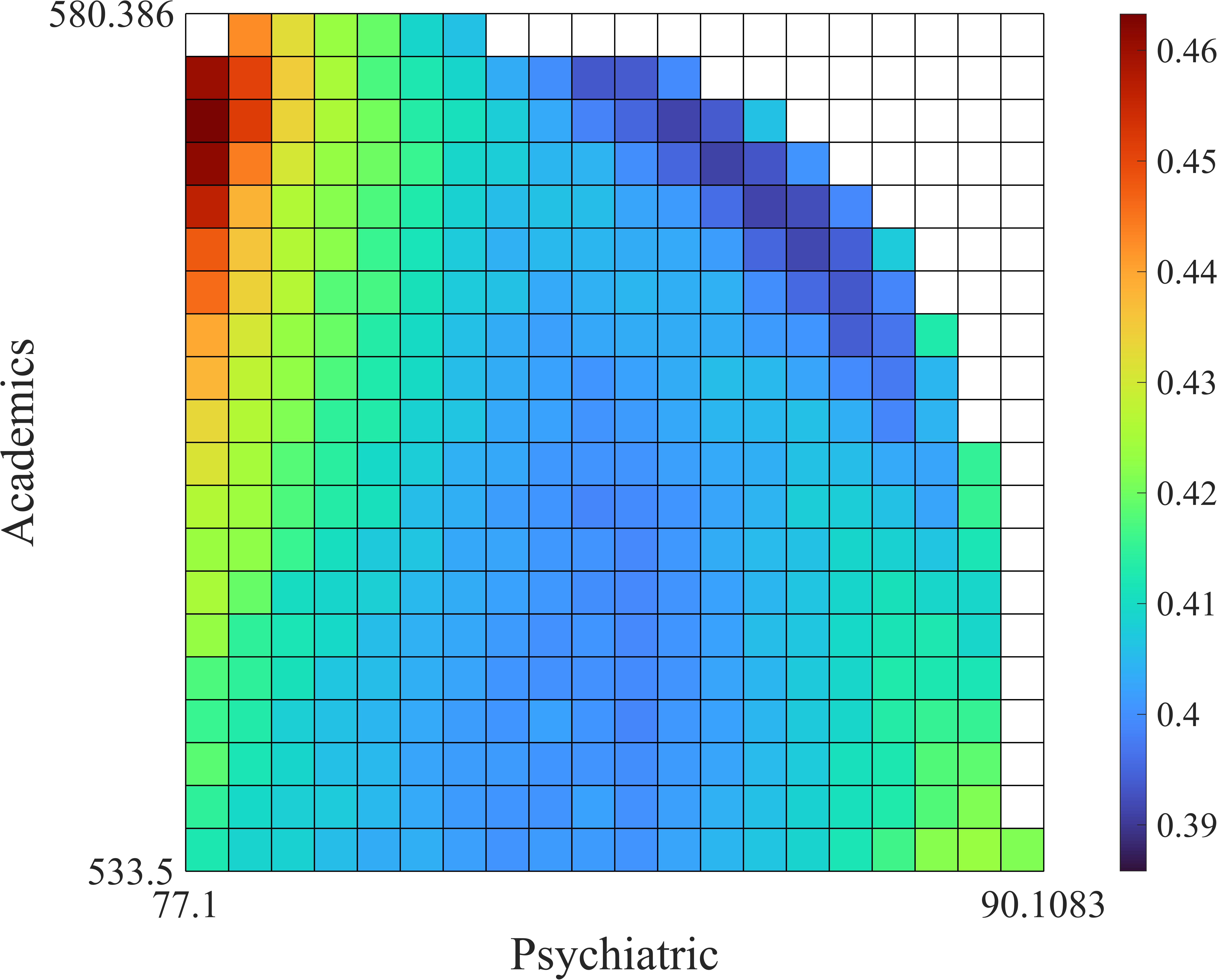}   
          \caption{Obj=BF, U(x)}
         \label{fig:BF-U-50}
    \end{subfigure}  
\begin{subfigure}[t]{0.24\textwidth}
        \centering
      \includegraphics[width=0.98\textwidth]{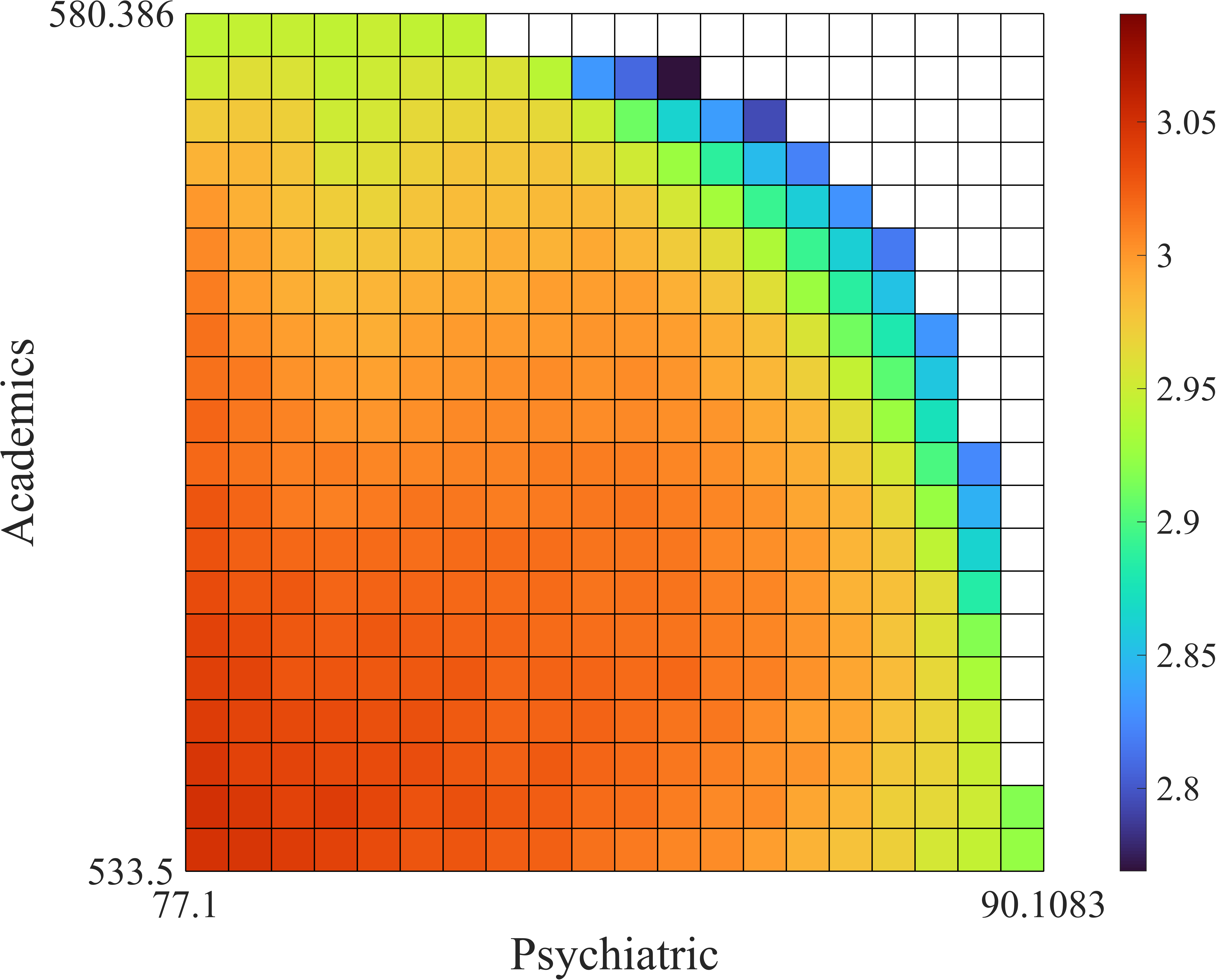}   
          \caption{Obj=BF, E(x)}
         \label{fig:BF-E-50}
    \end{subfigure}  
\begin{subfigure}[t]{0.24\textwidth}
        \centering
         \includegraphics[width=0.98\textwidth]{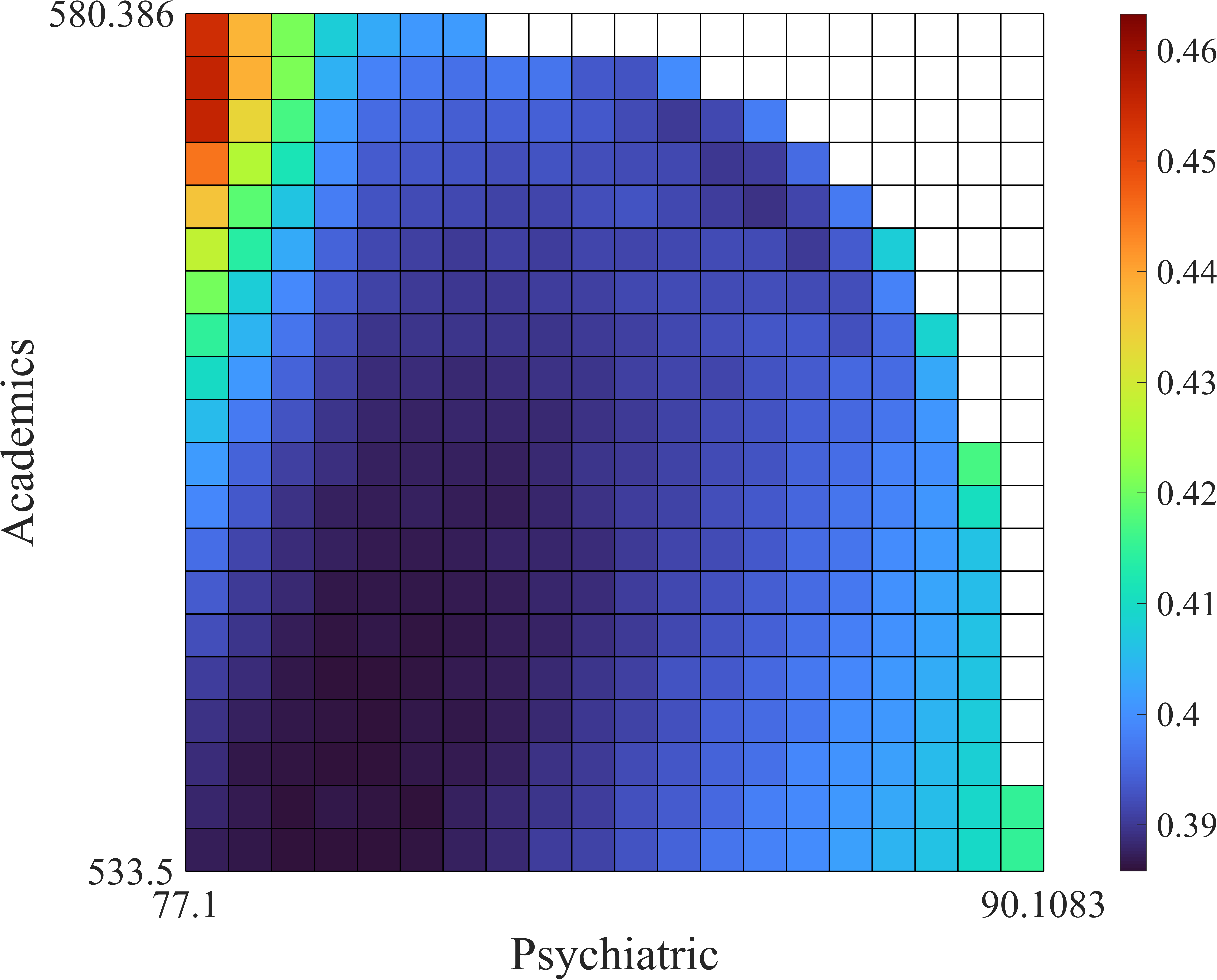}   
          \caption{Obj=BF, U(x)}
         \label{fig:BF-U-10}
    \end{subfigure}  %
 \begin{subfigure}[t]{0.24\textwidth}
        \centering
           \includegraphics[width=0.98\textwidth]{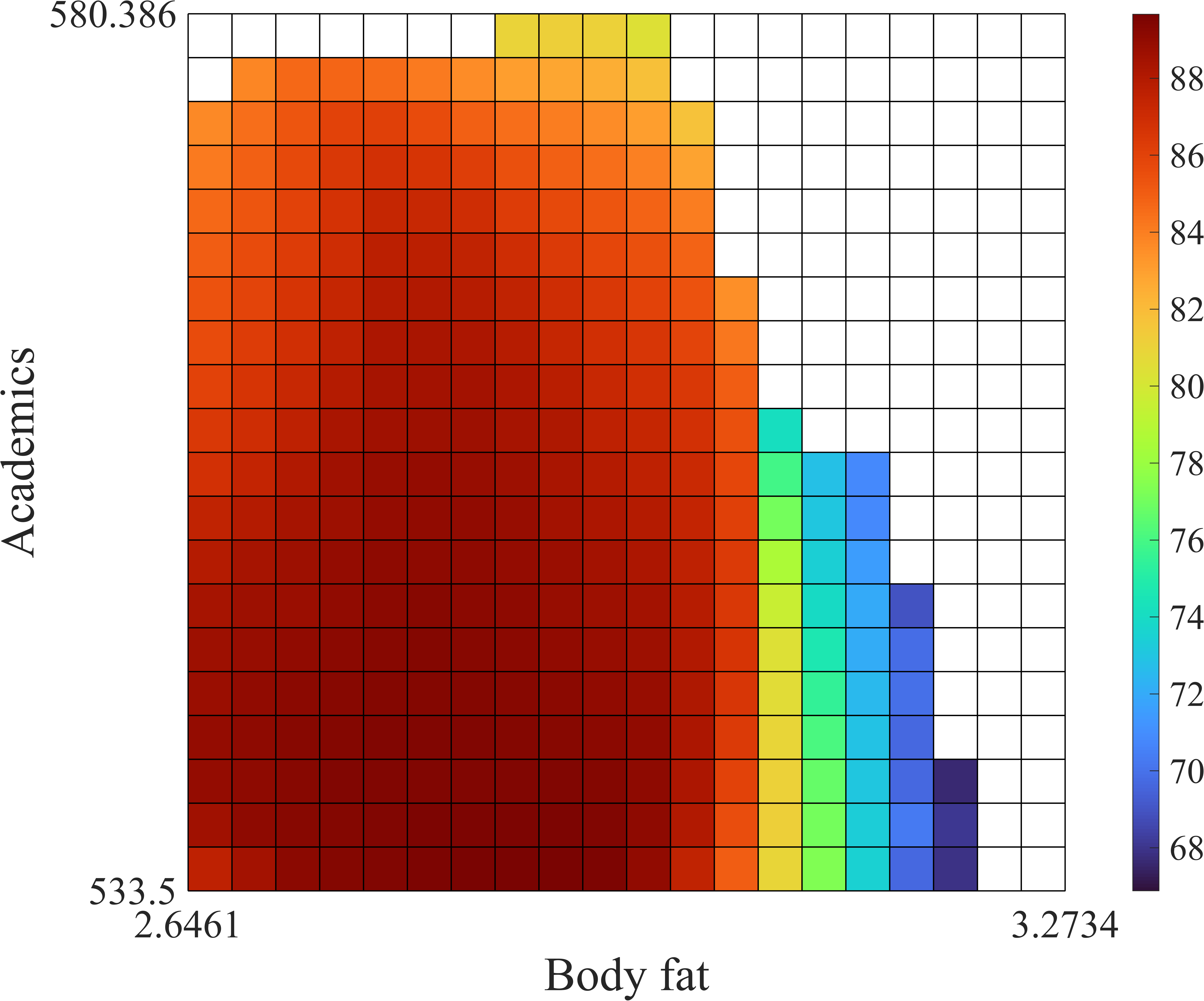}     
          \caption{Obj=Psy, E(x)}
         \label{fig:Psy-E-50}
    \end{subfigure} 
\begin{subfigure}[t]{0.24\textwidth}
        \centering
           \includegraphics[width=0.98\textwidth]{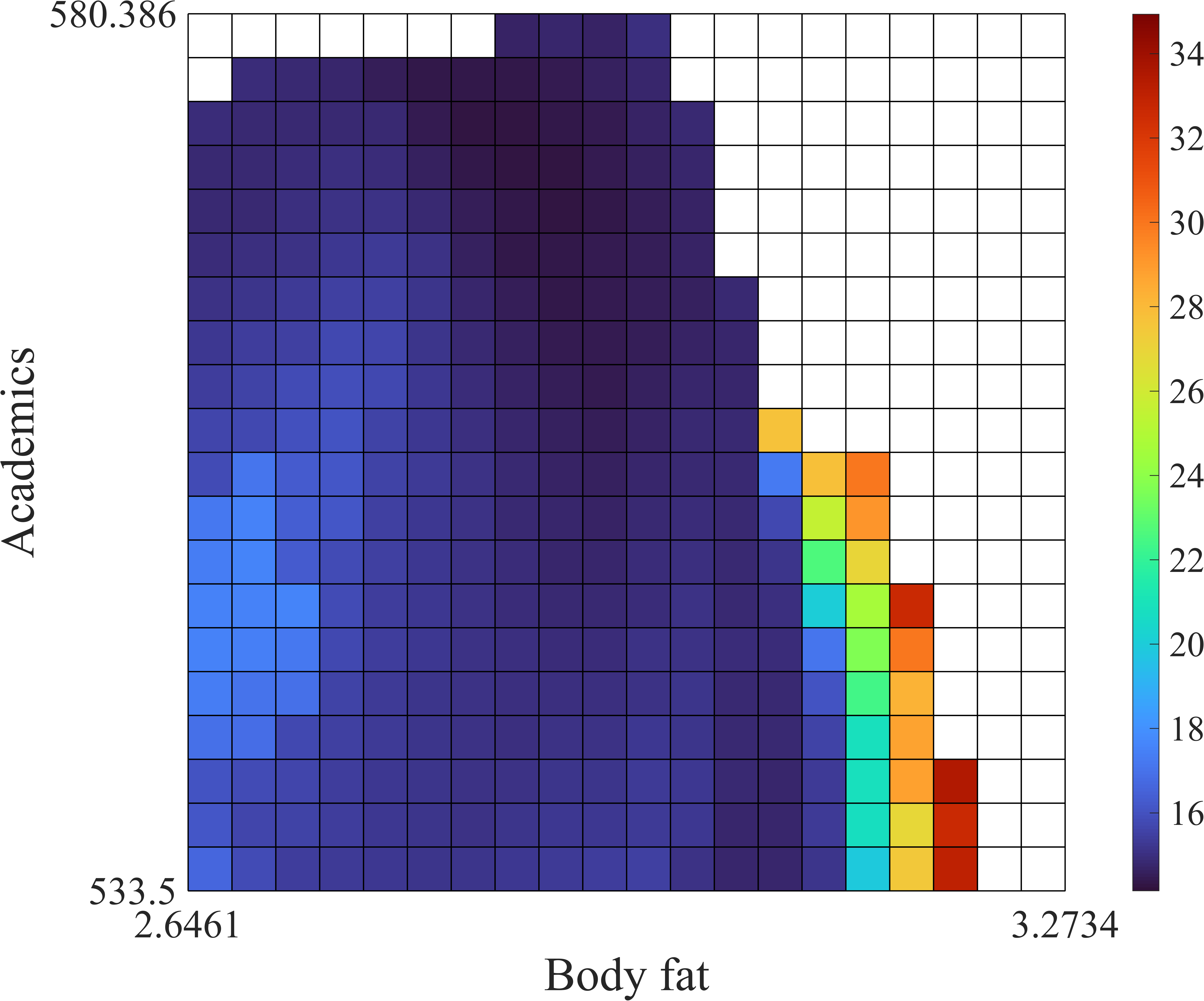}  
          \caption{Obj=Psy, U(x)}
         \label{fig:Psy-U-50}
    \end{subfigure} 
  \begin{subfigure}[t]{0.24\textwidth}
        \centering
             \includegraphics[width=0.98\textwidth]{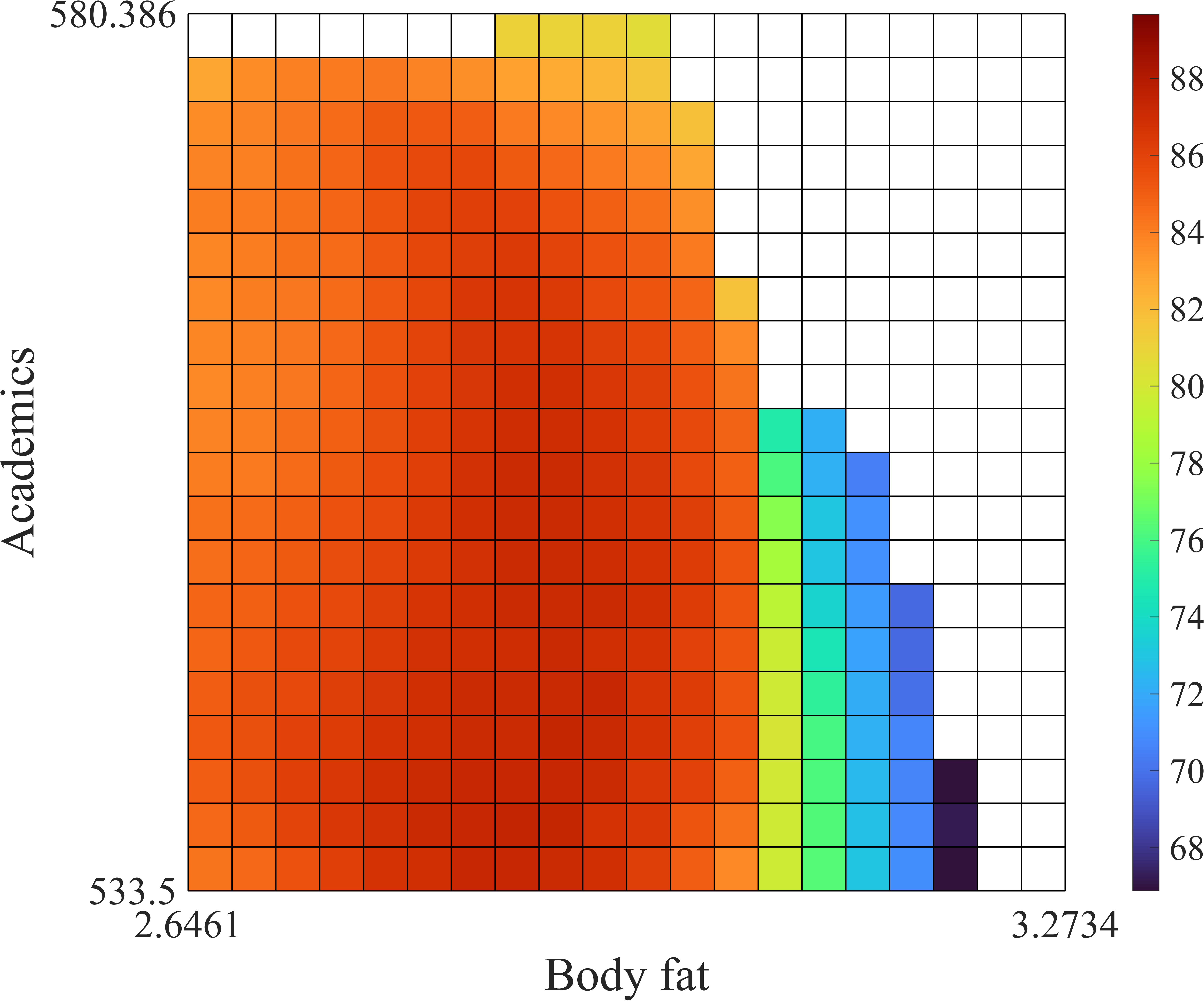}   
          \caption{Obj=Psy, E(x)}
         \label{fig:Psy-E-10}
    \end{subfigure} 
\begin{subfigure}[t]{0.24\textwidth}
        \centering
          \includegraphics[width=0.98\textwidth]{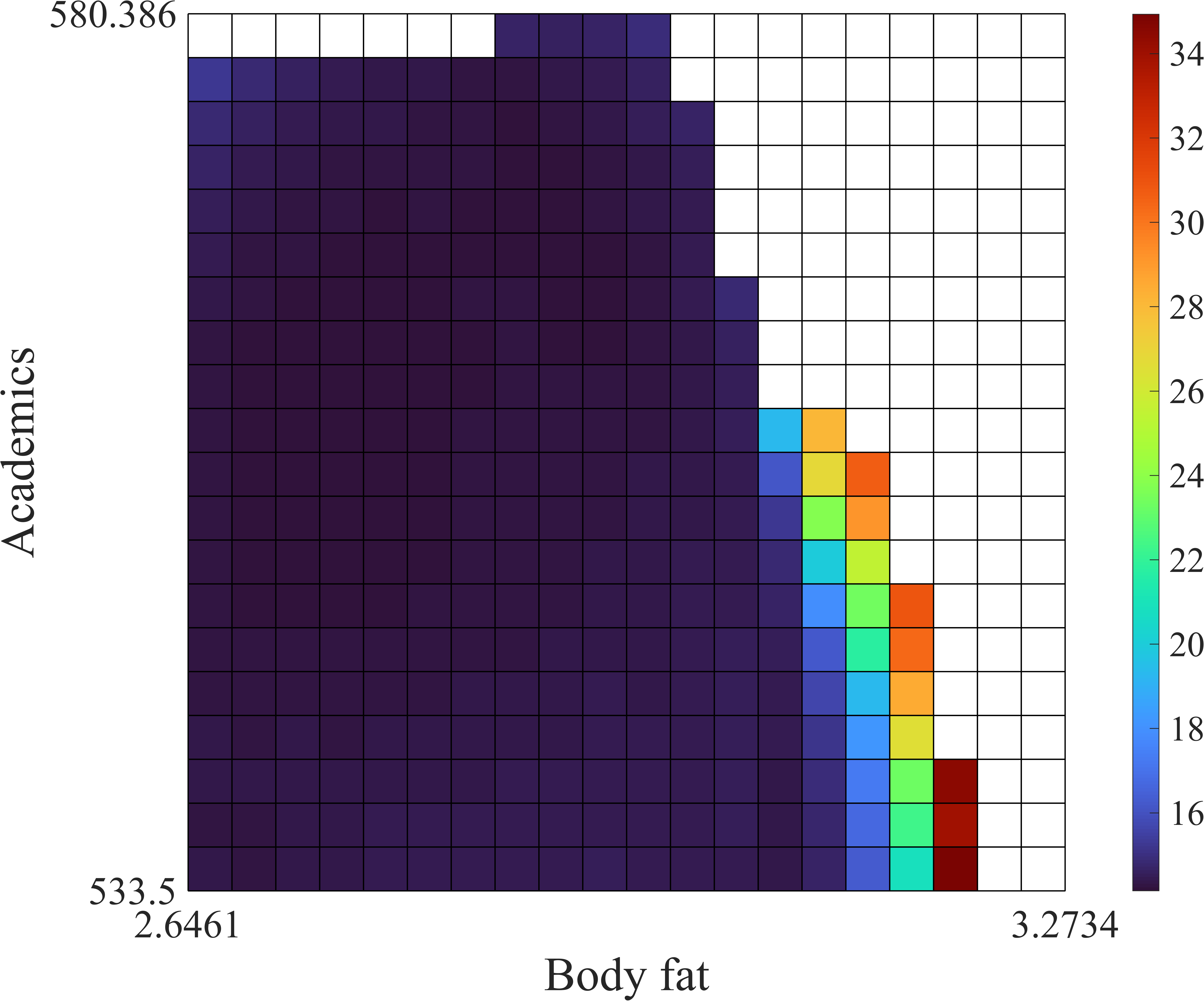}  
          \caption{Obj=Psy, U(x)}
         \label{fig:Psy-U-10}
    \end{subfigure} 
    \caption{
    Distribution of expected health benefits and associated uncertainties for the 7D model in OBS spaces using the standard approach (first two columns) and discounted approach with $C_{\alpha}=6.36134$ (last two columns).
    } 
    \label{fig:OBS_7} 
\end{figure}

\subsubsection{The objective-based behavioral space} 
We define the behavioral space using objective functions, selecting two objectives as descriptors and one as fitness. Descriptor ranges are based on maximum values from prior experiments, except body fat, which follows recommended values for 12-year-old [$14.1$, $26.4$]. Academic and psychiatric ranges are [$533.5$, $580.39$] and [$77.1$, $90.11$], with lower bounds set to reported mean values~\cite{dumuid2022yourbestday}. Figure~\ref{fig:OBS_7} presents the MAP-Elites results for the $7D$ time-use optimization problem in terms of expected values $E(x)$ and uncertainty $U(x)$ for different behavioral spaces and objectives, and compares the standard QD approach with our discounted approach.
For all objectives, the standard approach produces high-performing regions with maximum values of $580$ for academics, $2.6$ for body fat, and $90$ for psychiatric. These solutions cover a large portion of the behavioral space but extend into extreme regions (e.g., very low body fat or very high psychiatric scores).
In contrast, our approach leads to a more compact distribution of high-quality solutions. While maintaining comparable best performance, it removes extreme regions and concentrates solutions in more balanced areas (e.g., academics ($550$-$575$), body fat ($2.7$-$2.9$)), psychiatric ($75$-$85$), avoiding extreme and less reliable combinations.
In a nutshell, incorporating uncertainty leads to more constrained and realistic solution spaces, showing that some high-quality solutions under the standard approach are likely to high.
The uncertainty plots show a consistent pattern across all objectives. Under the standard approach, high-performing regions often coincide with elevated uncertainty, particularly near the boundaries of the behavioral space. For example, uncertainty values range for academics ($0.38$-$0.42$), body fat ($0.40$-$0.46$), and psychiatric ($0.30$-$0.34$). These results indicate that the highest uncertainty is observed for extreme combinations of outcomes.
With the discounted approach, solutions shift toward lower-uncertainty regions, avoiding unreliable extremes. Solutions are concentrated in central regions with lower uncertainty, indicating stronger support from real-world data.

\section{Conclusions}
We introduced a new QD framework for data-driven 24-hour time-use optimization that embeds predictive uncertainty directly into the search process. Using compositional data analysis to define health objectives and quantify uncertainty within each behavioral simplex, our approach explicitly balances expected outcomes with model confidence.
The results demonstrate that considering uncertainty shifts the recommended time-use patterns and reveals a clear divergence between regions of high expected benefit and regions of low uncertainty. Furthermore, our discounting approach leads to a substantial reduction in high-uncertainty elites, effectively guiding  the search toward more reliable regions of the high-dimensional behavioral space. 
The analysis of variable-based and objective-based behavioral representations highlights how uncertainty influences the distribution and quality of solutions. This reliable data driven time-use QD framework provides insight into the interaction between health benefits and reliability to find diverse, high-quality daily schedules.
A diverse solution set provides decision-makers with a wider range of time-use options that remain effective under different requirements and are robust to uncertainty. Consequently, embedding uncertainty into the QD process leads to more reliable data-driven time-use recommendations. This supports more robust and trustworthy decision-making in behavioral health optimization.

%\subsubsection{\ackname} 

\section*{Acknowledgments} This work has been supported by the National Health and Medical Research Council (NHMRC) through grant 2039039 and by the Australian Research Council (ARC) through grant FT200100536.

\bibliographystyle{abbrv}
\bibliography{qd,ref}

%\appendix

\end{document}